\documentclass{article}

\usepackage{color,xcolor}
\usepackage{epsfig}
\usepackage{graphicx}

\usepackage{adjustbox}
\usepackage{array}
\usepackage{booktabs}
\usepackage{colortbl}
\usepackage{float,wrapfig}
\usepackage{hhline}
\usepackage{multirow}
\usepackage{tabularx}
\usepackage{makecell}
\usepackage{float}

\usepackage{amsmath,amsfonts,amsthm,amssymb}
\usepackage{bm}
\usepackage{nicefrac}
\usepackage{microtype}

\usepackage{changepage}
\usepackage{extramarks}
\usepackage{fancyhdr}
\usepackage{lastpage}
\usepackage{setspace}
\usepackage{soul}
\usepackage{xspace}

\usepackage[pagebackref=true,breaklinks=true,colorlinks,citecolor={green},urlcolor={purple}]{hyperref}
\usepackage[nocompress]{cite}
\usepackage{url}

\usepackage{algorithm, algorithmic}
\usepackage{enumerate}
\usepackage{enumitem}
\usepackage{todonotes}

\usepackage{etoolbox,siunitx}
\usepackage{afterpage}
\usepackage{subcaption}
\usepackage[title]{appendix}
\usepackage{overpic}

\newcommand{\fid}{Fr\'echet Inception Distance\xspace}

\newcommand{\sect}[1]{Sec.~\ref{#1}}

\newcommand{\fig}[1]{Fig.~\ref{#1}}

\newcommand{\ignorethis}[1]{}

\makeatletter
\DeclareRobustCommand\onedot{\futurelet\@let@token\@onedot}
\def\@onedot{\ifx\@let@token.\else.\null\fi\xspace}

\def\eg{\emph{e.g}\onedot} 
\def\ie{\emph{i.e}\onedot}

\makeatother

\newcommand{\vv}[1]{\boldsymbol{#1}}
\newrobustcmd{\B}{\bfseries}
\newcommand*{\rom}[1]{\expandafter\romannumeral #1}
\newcommand{\Ns}{\mathcal{N}}

\definecolor{mydarkblue}{rgb}{0,0.08,1}
\definecolor{mydarkgreen}{rgb}{0.02,0.6,0.02}
\definecolor{mydarkred}{rgb}{0.8,0.02,0.02}
\definecolor{mydarkorange}{rgb}{0.40,0.2,0.02}
\definecolor{mypurple}{RGB}{111,0,255}
\definecolor{myred}{rgb}{1.0,0.0,0.0}
\definecolor{mygold}{rgb}{0.75,0.6,0.12}
\definecolor{myblue}{rgb}{0,0.2,0.8}
\definecolor{mydarkgray}{rgb}{0.66,0.66,0.66}

\newcommand{\myparagraph}[1]{\vspace{-6pt}\paragraph{#1}}

\usepackage{amsmath,amsfonts,bm}

\def\eqref#1{equation~\ref{#1}}
\def\1{\bm{1}}

\def\vv{{\bm{v}}}
\def\vw{{\bm{w}}}

\def\vz{{\bm{z}}}

\def\mC{{\bm{C}}}

\def\mF{{\bm{F}}}

\DeclareMathAlphabet{\mathsfit}{\encodingdefault}{\sfdefault}{m}{sl}
\SetMathAlphabet{\mathsfit}{bold}{\encodingdefault}{\sfdefault}{bx}{n}

\def\gW{{\mathcal{W}}}
\def\gX{{\mathcal{X}}}

\newcommand{\E}{\mathbb{E}}
\newcommand{\Ls}{\mathcal{L}}

    \usepackage[nonatbib,final]{neurips_2022}
\usepackage{bbding}

\title{Towards Diverse and Faithful One-shot Adaption of Generative Adversarial Networks}
\author{%
Yabo Zhang$^1$ \ Mingshuai Yao$^1$ \ Yuxiang Wei$^1$ \ Zhilong Ji$^2$ \ Jinfeng Bai$^2$ \ Wangmeng Zuo$^1$ $^{(}$\Envelope$^)$ 
\\
\\
\textsuperscript{1}Harbin Institute of Technology  \qquad
\textsuperscript{2}Tomorrow Advancing Life \\
}

\begin{document}

\maketitle

\begin{abstract}
One-shot generative domain adaption aims to transfer a pre-trained generator on one domain to a new domain using one reference image only.
However, it remains very challenging for the adapted generator (i) to generate diverse images inherited from the pre-trained generator while (ii) faithfully acquiring the domain-specific attributes and styles of the reference image. 
In this paper, we present a novel one-shot generative domain adaption method, \ie, DiFa, for diverse generation and faithful adaptation.
For global-level adaptation, we leverage the difference between the CLIP embedding of reference image and the mean embedding of source images to constrain the target generator. 
For local-level adaptation, we introduce an attentive style loss which aligns each intermediate token of adapted image with its corresponding token of the reference image.
To facilitate diverse generation, selective cross-domain consistency is introduced to select and retain the domain-sharing attributes in the editing latent $\mathcal{W}+$ space to inherit the diversity of pre-trained generator. 
Extensive experiments show that our method outperforms the state-of-the-arts both quantitatively and qualitatively, especially for the cases of large domain gaps.
Moreover, our DiFa can easily be extended to zero-shot generative domain adaption with appealing results.
Code is available at \url{https://github.com/1170300521/DiFa}.
%Recent works attempt to adapt a pre-trained generator using the power of CLIP models, yet failing to synthesis style-fidelity and diverse images under such an extreme setting.
%To combat it, we introduce two novel losses to better trade off between acquiring visual styles from the reference and inheriting the diversity from the original generator.
%the Attentive Perceptual loss attentively aligns each part of adapted images with its corresponding style in the reference.
%In the meanwhile, the Selective Cross-domain Consistency loss adaptively selects the common attributes, along with preserving them to inherit the prior knowledge.
\end{abstract}

\section{Introduction}
\label{introduction}
\begin{figure}[t]
   \begin{center}
   \includegraphics[width=.99\linewidth]{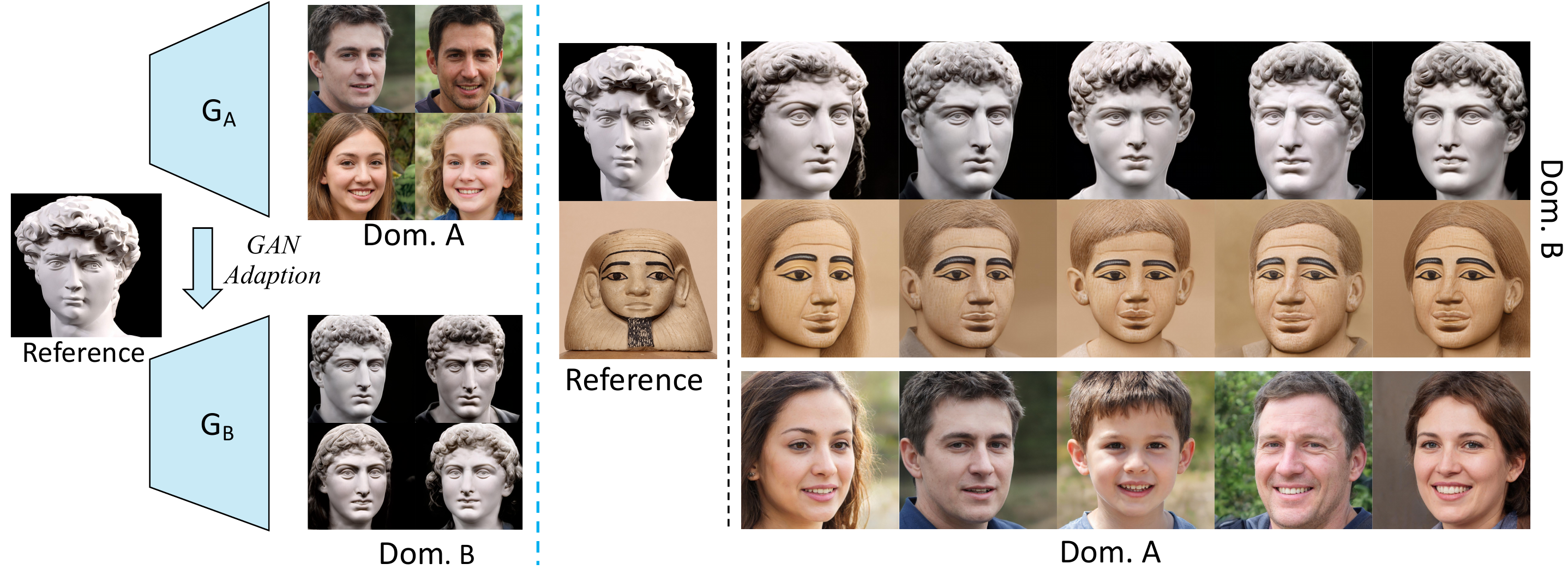}
   \end{center}
%   \vspace{-0.5em}
   \caption{\textbf{Diverse and faithful one-shot generative domain adaption.} 
   \textbf{Left: }One-shot generative domain adaption aims to transfer a pre-trained GAN from domain $A$ (\eg, FFHQ) to domain $B$ (\eg, sculptures) by providing one reference image only.
   \textbf{Right: }Synthesised images by our DiFa. 
   Notably, our DiFa is effective in inheriting the diverse generation ability of GAN from domain $A$, while faithfully acquiring the representative characteristics of the reference image in domain $B$.} 
    \label{fig:intro}
    \vspace{-1em}
\end{figure}

Generative adversarial networks (GANs)~\cite{goodfellow2014generative} have achieved remarkable progress in generating photo-realistic and highly-diverse images~\cite{karras2019style, karras2020analyzing,liu2022pretrainedGANs}. 
However, GANs usually require a large number of samples for stable training, and suffer from severe mode collapse when trained with insufficient data (\eg, one image).
Recently, several works~\cite{shaham2019singan,shocher2019ingan,bensadoun2021meta,zhang2018pa,zhao2020image,karras2020training,yang2021data} have been proposed to train a GAN from scratch with only one or few images, but are limited in generating high quality and diverse images.
%yet the diversity of synthesised images is limited and far from the real world (\eg, expression changes in faces). 
%
In this paper, we resort to one-shot or few-shot generative domain adaption (GDA), \ie, transferring a pre-trained generator on one domain to a new domain using one or few reference images (as shown in Fig.~\ref{fig:intro}). 
Thus, GDA can provide a new perspective to address the above issues by inheriting the generation ability and diversity of the pre-trained generator.

Many methods~\cite{wang2018transferring,mo2020freeze,robb2020few,yang2021one,li2020few,ojha2021few,zhao2022closer,xiao2022few} have been proposed for one-shot GDA.
%which usually adopt the adversarial loss with different training strategies. 
%
Nonetheless, domain-specific attributes and styles usually can be described by language, and thus can be well depicted by Contrastive-Language-Image-Pretraining (CLIP)~\cite{radford2021learning}.
%
%Recently, with the success of CLIP~\cite{radford2021learning},
%
Hence, CLIP-based one-shot GDA methods~\cite{gal2021stylegannada,zhu2021mind,kwon2022one,wang2022ctlgan} have been proposed to adapt a pre-trained generator, \eg, StyleGAN2~\cite{karras2020analyzing}, to the target domain. 
In particular, the domain-gap direction between source and target domains is first calculated in the CLIP embedding space. 
Then, the pre-trained generator is transferred by aligning the CLIP direction between the source and adapted images with the domain-gap direction. 
%
%Benefiting from the power of CLIP, the adapted GANs can generate photo-realistic reference domain images.

However, it remains very challenging for the adapted generator (i) to generate images as \emph{diverse} as the pre-trained generator while (ii) \emph{faithfully} acquiring the domain-specific attributes and styles of the reference image. 
Firstly, although the domain-gap direction extracts the pronounced characteristics of the reference image, the detailed local styles and attributes are usually ignored in CLIP embedding. 
Without considering these local styles and attributes, the adapted generator cannot faithfully acquire the domain-specific characteristics of the reference image.
Secondly, the domain-gap direction is the difference between the CLIP embedding of the reference image and source domain, which contains both domain-specific attribute shifts (\eg, thick eyebrows in second row of \fig{fig:intro}) and domain-sharing attribute shifts (\eg, gender).
Directly aligning the training sample-shift direction with the domain-gap direction introduces the unnecessary domain-sharing attribute changes to the adapted images, thereby being harmful to inheriting the diversity from the pre-trained generator. 
Although \cite{zhu2021mind,kwon2022one} proposed to use the style mixing and editing direction preservation to address these issues, only limited improvements are achieved.

In this work, we present a novel one-shot GDA method, \ie, DiFa, for diverse generation and faithful adaption.
In terms of faithful adaption, we consider both attributes and styles.
For global-level adaptation, we define the domain-gap direction as the difference between the CLIP embedding of reference image and the mean embedding of source images.
As for local-level adaptation, we introduce an attentive style (AS) loss on the intermediate layer of the CLIP image encoder. 
For each intermediate token of an adapted image, it first finds the nearest token of the reference image, and then minimizes their difference to make GDA adapt to the target style. 
%
%Specifically, we propose the Attentive Style Loss (AS) and Selective Cross-domain Consistency Loss (SCC) to address two issues, respectively. 
%
%Firstly, to faithfully extract the domain-specific styles of reference image, AS attentively aligns each intermediate style token of adapted image with its corresponding token of the reference image.
%
%Particularly, pre-trained CLIP image encoder is adopted as the feature extractor, and all tokens are extracted from the intermediate layer of the CLIP image encoder.
%
In terms of diverse generation, selective cross-domain consistency (SCC) is introduced to select and retain domain-sharing attributes in the editing latent $\mathcal{W}+$ space to inherit the diversity of pre-trained generator. 
%
%Secondly, to inherit the diversity from the pre-trained generator, SCC adaptively selects the domain-sharing attributes in the $\mathcal{W}+$ space, and retains these attributes when adapting. 
%
In particular, we use a styleGAN inversion models~\cite{richardson2021encoding, tov2021designing} to invert the images from source and target domains into the $\mathcal{W}+$ space.
Then, we compute the direction $\Delta \vw$ between the two domains, where smaller values in $\Delta \vw$ indicate that the corresponding latent variables in $\mathcal{W}+$ space are domain-sharing attributes.
%
%Lower value in $\Delta \vw$ means less unique in the reference domain, and corresponds to the domain-sharing attribute. 
%
Selective cross-domain consistency encourages an adapted image and its corresponding source image to be similar in domain-sharing attributes, and can be different in other attributes.  
SCC allows the adapted generator to inherit from the pre-trained generator selectively. 
Thus our DiFa can guarantee the diversity of adapted images without the sacrifice of decreasing domain adaption ability. 

%Thus, we constrain the low value dimensions to be unchanged when adapting, and encourage the adapted generator to generate diverse reference domain images.

Quantitative and qualitative experiments are conducted on a wide range of source and target domains.
Evaluation results highlight the superiority of our DiFa compared against the state-of-the-art methods, especially for the cases of large domain gaps (\eg, Cat $\to$ Tiger).
To illustrate the editing capabilities of the adapted latent space, we
employ InterFaceGAN~\cite{shen2020interfacegan} to edit the real images in target domain.
%
% Moreover, with minor modifications, our DiFa can easily be extended to zero-shot GDA with compelling results.

% \vspace{-1em}
% Overall, our contributions are summarized as follows: %
% (1) a novel method DiFa for one-shot generative domain adaption,
% (2) the attentive style and selective cross-domain consistency losses for diverse generation and faithful adaption,
% and (3) the extension to the zero-shot setting with appealing results.

Overall, our contributions are summarized as follows: %
\vspace{-3.5mm}
\begin{itemize}[itemsep=0pt, parsep=0pt, leftmargin=10pt]
    \item We introduce a novel method namely DiFa, along with selective cross-domain consistency and attentive style losses, for diverse generation and faithful adaption.
    \item Extensive experiments show the effectiveness of our DiFa in acquiring the representative domain characteristics from the reference image, and inheriting the ability of pre-trained generator to produce diverse images.
    \item Our DiFa can be easily extended to zero-shot generative domain adaption with appealing results.
\end{itemize}

\section{Related Work}
\vspace{-0.3em}
\myparagraph{Few-shot Domain Adaption of GANs.}
Few-shot generative domain adaption aims to transfer a generator pre-trained on a source domain to a new target domain with very limited reference images.
Earlier studies~\cite{wang2018transferring,mo2020freeze,robb2020few,yang2021one,li2020few,ojha2021few,zhao2022closer,xiao2022few} utilized the adversarial loss~\cite{goodfellow2014generative} to capture domain-specific information from given reference images.
To reduce mode collapse, these methods usually adopted fewer learnable parameters~\cite{wang2018transferring,mo2020freeze,robb2020few,yang2021one} or introduce regularization terms~\cite{li2020few,ojha2021few,zhao2022closer,xiao2022few}, but still produce images with insufficient diversity.
With the success of CLIP~\cite{radford2021learning}, recent works~\cite{gal2021stylegannada,zhu2021mind,kwon2022one,wang2022ctlgan} leveraged the difference between the CLIP embeddings of the source and target domains to guide the attribute-level adaption, beating the methods with adversarial loss~\cite{goodfellow2014generative}.
To better capture domain-specific styles, several methods~\cite{zhu2021mind,kwon2022one} adopted the style mixing trick during inference time, however, it may bring undesired semantic artifacts when there is a significant shape discrepancy.
~\cite{zhu2021mind,kwon2022one,wang2022ctlgan} attempted to generate diverse images by preserving the editing distance of input pairs, before and after adaption.
Nonetheless, they indistinguishably retain both domain-sharing and domain-specific attributes, which is conflicted with faithful adaption.

\myparagraph{GAN Inversion.}
GAN inversion aims to invert an image into its corresponding latent codes, which can be grouped into optimization-based and encoder-based methods.
Optimization-based inversion~\cite{zhu2016generative,creswell2018inverting} directly updates the latent code by minimizing the reconstruction error.
Albeit high-quality and accurate reconstruction can be obtained, it usually costs a few minutes for an image.
%
%and stops the gradients of input image.
In contrast, encoder-based algorithms~\cite{abdal2019image2stylegan,richardson2021encoding,tov2021designing} directly embed a given image into latent codes, so that the inference can be completed in real-time, and the gradients of input images could also be passed backward. 
Moreover, encoder-based algorithms also achieve considerable performance when handling out-of-domain images, and thus it is feasible to project adapted images into $\gW^+$ codes during training.
%
% ~\cite{shen2020interfacegan,abdal2021clip2stylegan,shen2021closed} have discovered many meaningful editing directions in $\gW$ space, and we intend to leverage them
% Note that there are many meaningful editing directions~\cite{shen2020interfacegan,abdal2021clip2stylegan,shen2021closed} which each just control one visual attribute in well-disentangled $\gW+$ space and any two editing directions are roughly independent of each other, hence, we could deduce that each of $\gW+$ dimensions manipulates one attribute at most.
% \subsection{CLIP-based Image Editing and Generation}

% \subsection{Few-shot image Generation}
\vspace{-0.5em}
\section{Proposed Method}
\vspace{-0.5em}
\label{method}

\begin{figure}[t]
   % \vspace{-1em}
   \begin{center}
   \includegraphics[width=.99\linewidth]{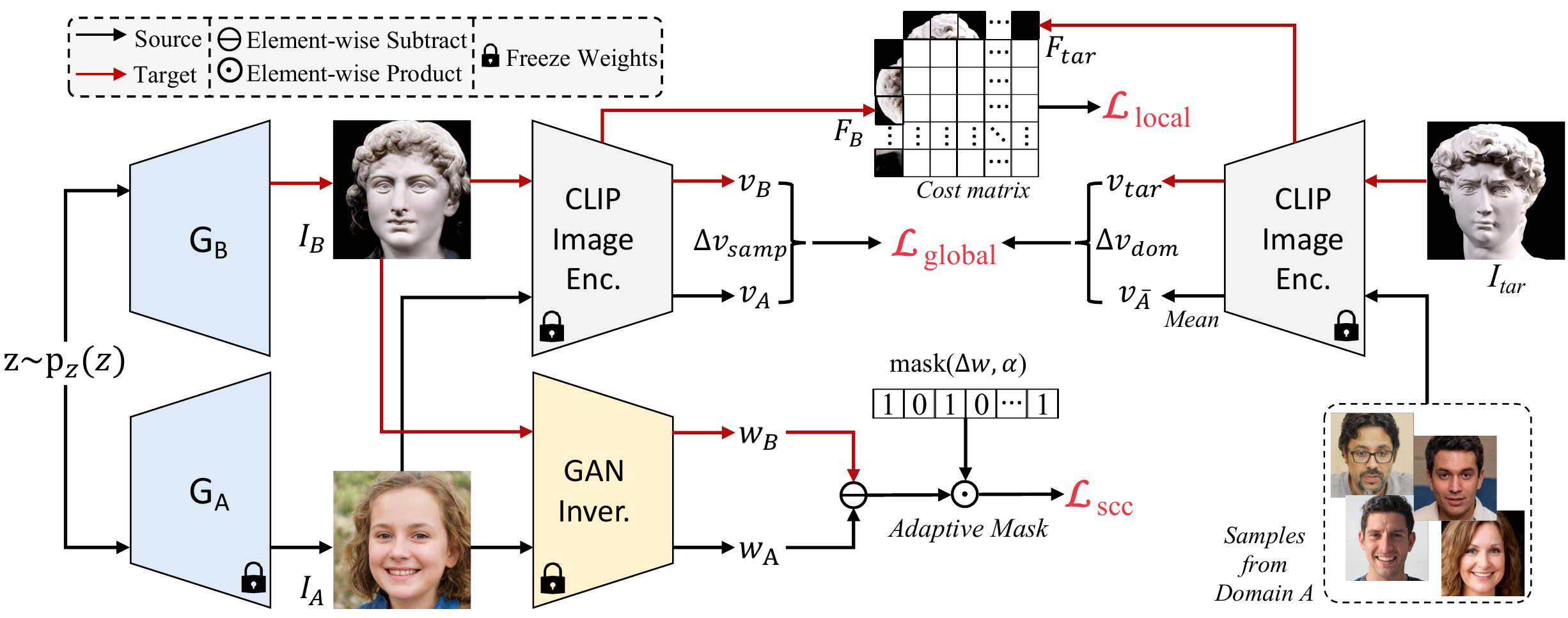}
   % \fbox{\rule{0pt}{2in} \rule{.9\linewidth}{0pt}} 
   \end{center}
   \vspace{-2mm}
   \caption{
   \textbf{Overview of our DiFa.}
   The adapted generator $G_B$ is initialized by pre-trained generator $G_A$.
   With the aid of CLIP image encoder, the global-level adaption loss $\Ls_{global}$ and attentive style loss $\Ls_{local}$ encourage $G_B$ to faithfully acquire both global and local representative domain-specific characteristics from the reference $I_{tar}$.
   To facilitate diverse generation inherited from $G_A$, the selective cross-domain consistency loss $\Ls_{scc}$ selects and retains domain-sharing attributes.}
    % with mask$(\Delta \vw,\alpha)$
    \label{fig:archetecture}
    \vspace{-4mm}
\end{figure}

In this work, we focus on one-shot generative domain adaption task, which aims to transfer a generator $G_A$ pre-trained on domain $A$ to a new domain $B$ using one reference image $I_{tar}$ only.
Specifically, we present a novel method termed DiFa to generate diverse images inherited from the pre-trained generator while faithfully acquiring the domain-specific attributes and styles of the reference image.
The overview of our DiFa is illustrated in \fig{fig:archetecture}.
In this section, we first introduce the global-level adaption loss with an estimated domain-gap direction. 
The attentive style loss and selective cross-domain consistency loss are then proposed for local-level adaptation and diverse generation, respectively.
Finally, we introduce the overall learning objective for training.

% In this work, we focus on adapting a generator $G_A$ pre-trained on domain A to a new domain B with one image $I_{tar}$ only.
% Initially, we introduce a preliminary to guide the attribute-level adaption with an estimated domain-direction (\sect{baseline}).
% According to this preliminary, the Attentive Style loss encourages $G_B$ to acquire domain-specific visual styles (\eg, colors and representative parts) from the reference (\sect{AP}), rather than leveraging the style mixing trick during inference time.
% Subsequently, the Selective Cross-domain Consistency loss adaptively selects and preserves domain-sharing attributes in the $\mathcal{W}+$ space, to inherit the diversity from the original generator (\sect{SCC}).
% Ultimately, \sect{overall} describes the overall loss function during training.
\vspace{-0.2em}
\subsection{Global-level Adaption}
\vspace{-0.2em}
\label{baseline}
Recent studies~\cite{gal2021stylegannada,zhu2021mind,kwon2022one} have demonstrated the superiority of CLIP in transferring a generator in one domain to a new domain under the one-shot setting.
In comparison to the methods based on adversarial loss~\cite{wang2018transferring,mo2020freeze,robb2020few,yang2021one,li2020few,ojha2021few,zhao2022closer,xiao2022few}, CLIP-based methods are effective in describing domain characteristics and resulting in photo-realistic images.
Given a generator $G_A$ pre-trained on domain $A$ and a target reference image $I_{tar}$ from domain $B$, CLIP-based methods first calculate the domain-gap direction between domain $B$ and $A$:
\vspace{-2mm}
\begin{equation}
\label{eq:d_domain}
\Delta \vv_{dom} = \vv_{tar} - \vv_{src},
\end{equation}
where $\vv_{tar} = E_I(I_{tar})$ denotes the embedding of target domain $B$ and $E_I$ is the CLIP image encoder. 
$\vv_{src}$ represents the CLIP embedding of source domain $A$.  
To transfer $G_A$ to domain $B$, they copy a new generator $G_B$ from $G_A$ and finetune it by aligning the sample-shift direction $\Delta \vv_{samp}$ with the domain-gap direction $\Delta \vv_{dom}$:
\begin{equation}
% \vspace{-2mm}
\label{eq:d_sample}
\Delta \vv_{samp} = \vv_{B} - \vv_{A},
\end{equation}
\begin{equation}
\label{eq:l_attr}
\Ls_{global} = 1 - \frac {\Delta \vv_{samp} \cdot \Delta \vv_{dom}} {|\Delta \vv_{samp}||\Delta \vv_{dom}|},
\end{equation}
where $\vv_{B} = E_I(G_B(\vz))$ and $\vv_{A} = E_I(G_A(\vz))$ denote the CLIP embeddings of domain $B$ and domain $A$ samples. 
$\vz\sim\Ns(0,I)$ denotes the input noise. 
After finetuning with the global-level adaption loss $\Ls_{global}$, the adapted generator $G_B$ can generate high-quality images of domain $B$.

Note that the embedding of source domain $\vv_{src}$ can be calculated in different ways. 
Two-stage methods~\cite{zhu2021mind,kwon2022one} find the image corresponding to $I_{tar}$ in domain $A$ and treat its CLIP-space embedding as $\vv_{src}$.
Nonetheless, the corresponding image in domain $A$ inevitably contains domain-specific attributes of $I_{tar}$, leading to $\Delta \vv_{dom}$ ignoring these domain-specific attributes. 
One-stage methods~\cite{gal2021stylegannada} utilize the mean embedding of source images as $\vv_{src}$.
Intuitively, the mean embedding usually represents the common attributes of source domain, and does not affect domain-specific attribute shifts in $\Delta \vv_{dom}$.
Thus, in our experiments, we use the mean embedding of source images as the source domain embedding, \ie, $\vv_{src} = \vv_{\bar A} = \E_{\vz\sim\Ns(0, I)}[E_I(G_A(\vz))]$.

\vspace{-0.5em}
\subsection{Local-level Adaption}
\vspace{-0.5em}
\label{AP}

Albeit the domain-gap direction $\Delta \vv_{dom}$ captures the global-level representative domain characteristics of reference image $I_{tar}$, the local attributes and visual styles are usually ignored in CLIP embedding. 
Therefore, training $G_B$ with $\Ls_{global}$ only cannot faithfully capture the local-level domain-specific characteristics of $I_{tar}$.
For example, images generated by StyleGAN-NADA~\cite{gal2021stylegannada} fail to acquire the mane and stripes of tigers during the Cat $\to$ Tiger adaption (the first row in \fig{fig:main_cat}(c)).
\cite{zhu2021mind,kwon2022one} tried to inherit the detailed visual styles from $I_{tar}$ through the style mixing.
However, when there is a significant shape discrepancy (\eg, pose or cross-category) between $I_{tar}$ and the original adapted image $I_B$, the mismatch of content and style in $I^{mix}_B$ will lead to visible artifacts (the last row in \fig{fig:main_ffhq}(b) and the first row in \fig{fig:main_cat}(b)).

To mitigate the above issue, we further present an attentive style loss $\Ls_{local}$ to help $G_B$ faithfully acquire the local-level representative attributes and styles of $I_{tar}$.
Inspired by content-style alignment in style transfer~\cite{kolkin2019style}, $\Ls_{local}$ is designed to encourage each part of $I_B$ to attentively align with its corresponding styles from $I_{tar}$.
Specifically, we first extract the intermediate tokens of $I_B$ and $I_{tar}$ from the $k$-th layer of CLIP image encoder (shown in Fig.~\ref{fig:attn_perc}), and then align each of adapted tokens $\mF_B$ with its closest target token from $\mF_{tar}$, where $\mF_B=\{\mF_B^1,\dots,\mF_B^n\}$ and $\mF_{tar}=\{\mF_{tar}^1,\dots,\mF_{tar}^m\}$ are the extracted tokens.
The final attentive style loss is defined as,

\vspace{-1em}
\begin{equation}
% \vspace{-2mm}
    \label{l_as}
    \Ls_{local} = \max \Big(\frac 1 n \sum_i \min_j \mC_{i,j}, \frac 1 m \sum_j \min_i \mC_{i,j} \Big),
% \end{align}
\end{equation}
where $\mC$ is the cost matrix to measure the token-wise distances from $F_B$ to $F_{tar}$, and each element of $\mC$ is computed as:
\begin{align}
\label{eq_cost}
    \mC_{i,j} = 1 - \frac {\mF_B^i \cdot \mF_{tar}^j} {|\mF_B^i| |\mF_{tar}^j|}.
\end{align}

\subsection{Selectively Diverse Generation}
\label{SCC}
\begin{figure}[t]
\centering
\begin{minipage}{.48\textwidth}
    \centering
    \includegraphics[width=0.99\linewidth]{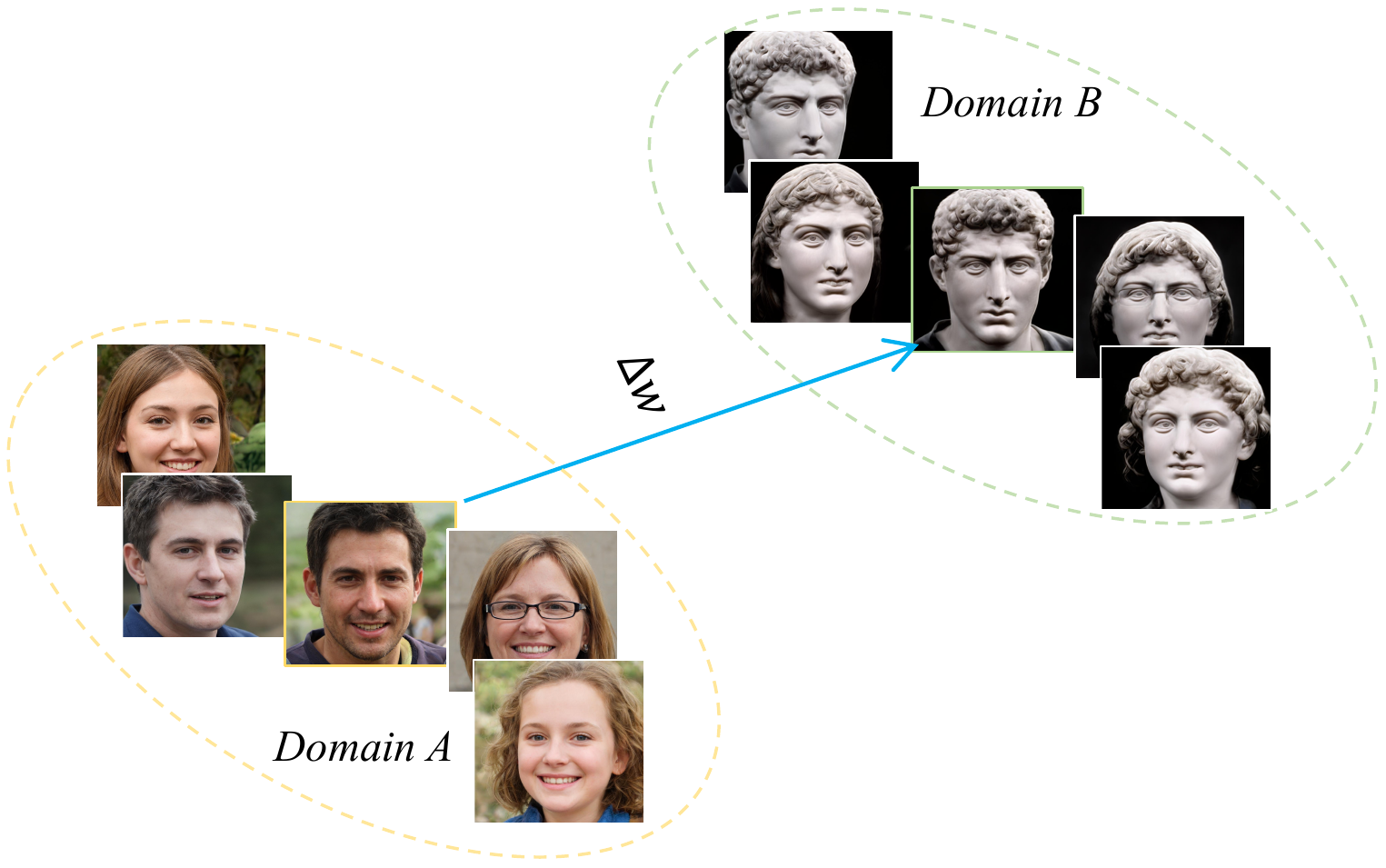}
    \vspace{-2em}
    \caption{
    \textbf{The process of computing $\Delta \vw$ between domain $A$ and domain $B$.}
    Given two clusters of $\gW+$ codes from domain $A$ and domain $B$, $\Delta \vw$ is defined as the difference of their cluster centers.
    }
    \label{fig:compute_w}
\end{minipage}
\hfill
\begin{minipage}{.48\textwidth}
    \centering
    \includegraphics[width=0.99\linewidth]{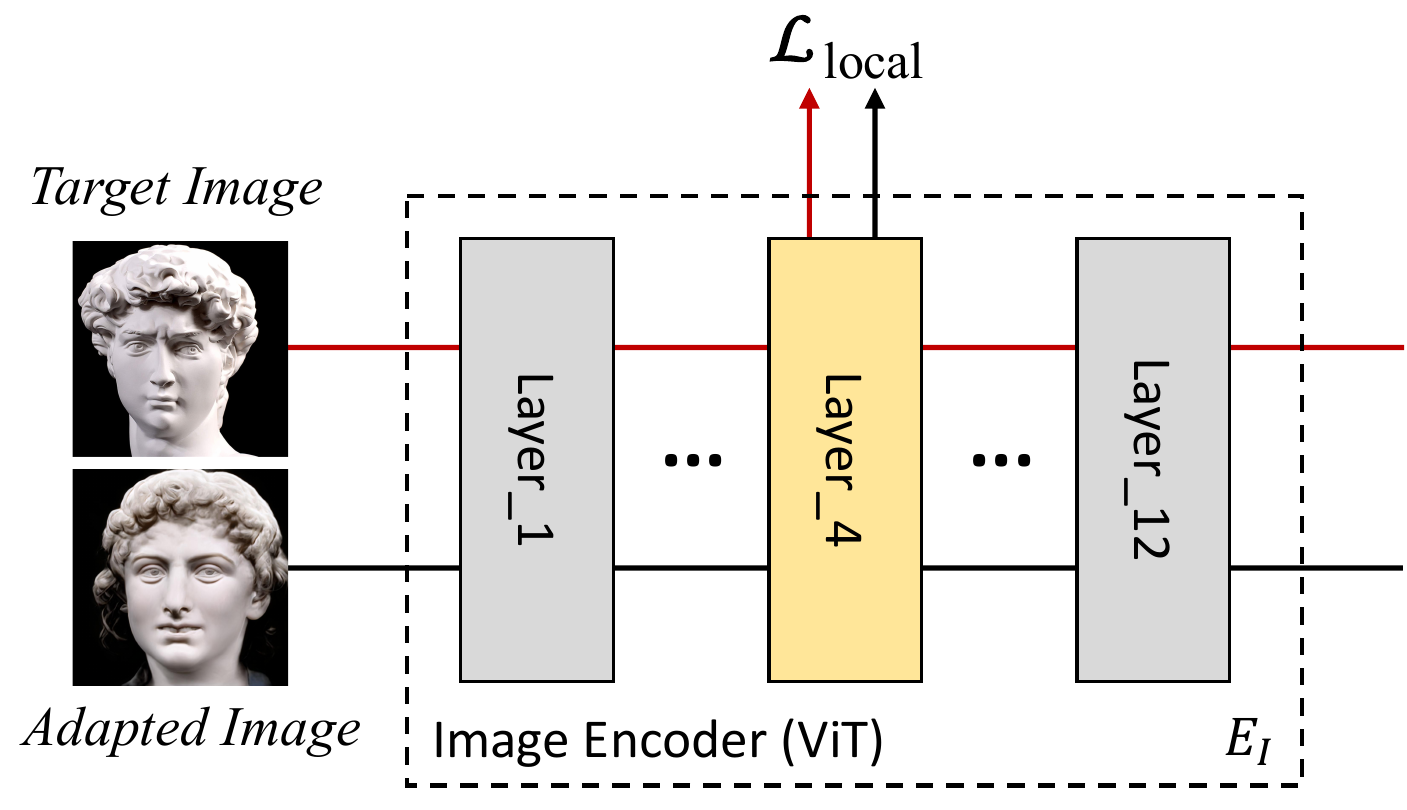}
    \caption{
    \textbf{Illustration of the Attentive Style loss.}
    We extract the intermediate adapted and target tokens from the Layer\_4 of CLIP image encoder, following by aligning each adapted token with its corresponding target token attentively.}
    \label{fig:attn_perc}
\end{minipage}
\vspace{-1.5em}
\end{figure}
The ability to generate diverse target domain images is also critical for one-shot generative domain adaption. 
Recall that the domain-gap direction is the difference between the embedding of $I_{tar}$ and source domain, which contains both domain-specific and domain-sharing attribute shifts.
Training $G_B$ with $\Ls_{local}$ also introduces the unnecessary domain-sharing attribute changes to the adapted images, which hinders $G_B$ from inheriting the diversity of the pre-trained generator $G_A$. 
To facilitate diverse generation, we propose a selective cross-domain consistency loss to select and retain the domain-sharing attributes in $\mathcal{W}+$ space.
Intuitively, if an attribute is similar between domains $A$ and $B$ during adaption, it is more likely to be a domain-sharing attribute.
According to this assumption, we can dynamically analyze and preserve the domain-sharing attributes.
Specifically, we first invert $G_A(\vz)$ and $G_B(\vz)$ into $\mathcal{W}+$ latent codes $\vw_A$ and $\vw_B$ with an pre-trained inversion model (\eg, pSp~\cite{richardson2021encoding} or e4e~\cite{tov2021designing}) for each iteration.
Then, as shown in Fig.~\ref{fig:compute_w}, we compute the difference $\Delta \vw$ between the centers of a queue of $\gW+$ latent codes $\gX_{A}$ and a queue of $\gW+$ latent codes $\gX_{B}$, where $\gX_A$ and $\gX_B$ are dynamically updated with $\vw_A$ and $\vw_B$ during training.
According to $\Delta \vw$, we encourage $\vw_A$ and $\vw_B$ to be consistent in channels with less difference,
\begin{equation}
    \label{eq4}
    \Ls_{scc} = ||\text{mask}(\Delta \vw, \alpha) \cdot (\vw_B - \vw_A)||_{1},
\end{equation}
where $\alpha$ represents the proportion of preserved attributes and mask($\Delta \vw, \alpha$) determines which channels to be retained. 
Let $|\Delta \vw_{s_{\alpha N}}|$ be the $\alpha N$-th largest element of $|\Delta \vw|$, and each dimension of mask($\Delta \vw, \alpha$) is calculated as:
\begin{equation}
    \label{eq5}
    \text{mask}(\Delta \vw, \alpha)_i=
    \begin{cases} 
    1& |\Delta \vw_i| < |\Delta \vw_{s_{\alpha N}}|\\
    0& |\Delta \vw_i| \ge |\Delta \vw_{s_{\alpha N}}|\\
    \end{cases}.
\end{equation}

Note that fine layers of StyleGAN~\cite{karras2019style, karras2020analyzing} usually control color information, and constraining them may have a detrimental effect on obtaining styles of $I_{tar}$. 
Hence, we only use latent codes corresponding to coarse spatial resolutions ($4^2– 8^2$) and middle resolutions ($16^2– 32^2$) in $\Ls_{scc}$.

\subsection{Overall Training Loss}
\label{overall}

Our overall training loss consists of three terms, \ie, the global-level adaption loss $\Ls_{global}$, the attentive style loss $\Ls_{local}$ for acquiring detailed style information and the selective cross-domain consistency loss $\Ls_{scc}$ for inheriting the diversity:
\begin{equation}
    \label{eq6}
    \Ls_{overall} = \Ls_{global} + \lambda_{local}\Ls_{local} + \lambda_{scc}\Ls_{scc}.
\end{equation}
In our experiments, we use $\lambda_{local} = 2$ and $\lambda_{scc}=\max(0, \frac {n_{iter} - n_B} {N_{iter} - n_{iter}})$, where $N_{iter}$ and $n_{iter}$ denote the total number of training iterations and the $n_{iter}$-th iteration of training, respectively.
That is, $\lambda_{scc}$ increases linearly as the training proceeds.

% Our overall training loss consists of three terms, \ie, the attribute-level loss $\Ls_{attr}$, the Attentive Style loss $\Ls_{as}$ for acquiring more sufficient style information and the Selective Cross-domain Consistency loss $\Ls_{scc}$ for inheriting the diversity:
% \begin{equation}
%     \label{eq6}
%     \Ls_{overall} = \Ls_{attr} + \lambda_{as}\Ls_{as} + \lambda_{scc}\Ls_{scc}.
% \end{equation}
% Empirically, $\lambda_{ap}$ is set to two and $\lambda_{scc}=\max(0, \frac {n_{iter} - n_B} {N_{iter} - n_{iter}})$, where $N_{iter}$ and $n_{iter}$ denote the total number of training iterations and the $n_{iter}$th iteration of training respectively.
% We increase $\lambda_{scc}$ linearly since $\Delta \vw$ becomes more accurate as the training proceeds.

\section{Experiments}
% \vspace{-0.5em}
\label{experiments}
\begin{figure}[t]
%   \vspace{-1em}
  \begin{center}
  \includegraphics[width=.99\linewidth]{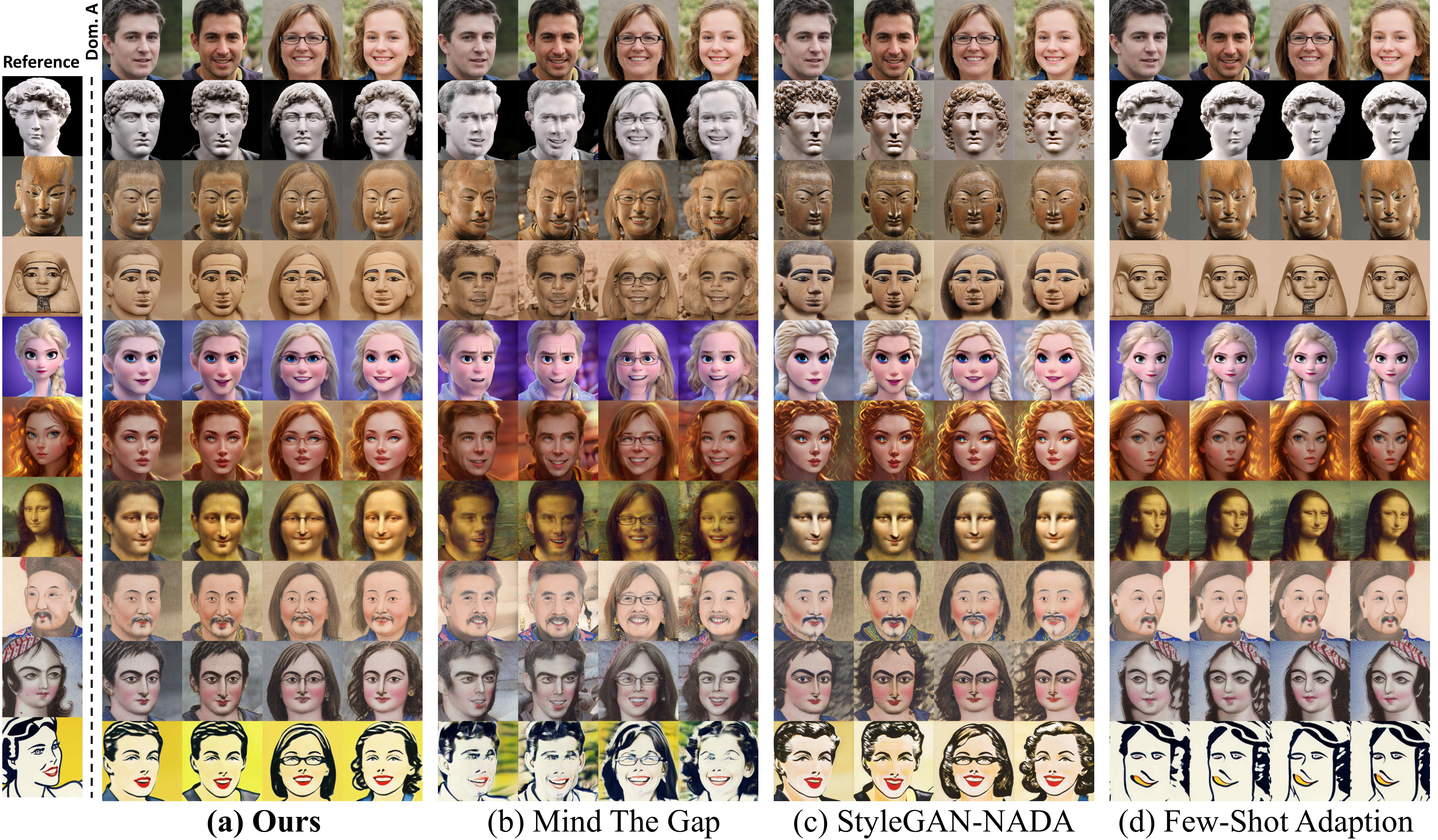}
  \end{center}
  \vspace{-1mm}
  \caption{
  \textbf{Qualitative comparisons using the generator pre-trained on FFHQ}~\cite{karras2019style} between our DiFa, Mind The Gap~\cite{zhu2021mind}, StyleGAN-NADA~\cite{gal2021stylegannada} and Few-Shot Adaption~\cite{ojha2021few}.
  The first row and first column show source images in domain $A$ and reference images in domain $B$.
  Our DiFa not only inherits the ability from the pre-trained generator to produce highly diverse and photo-realistic images, but also faithfully acquires the representative characteristics from the reference images, significantly outperforming the competing methods.
  \textbf{Results best seen at 500\% zoom.}
  }
    \label{fig:main_ffhq}
  \vspace{-1mm}
\end{figure}
\begin{figure}[t]
   % \vspace{-1em}
   \begin{center}
   \includegraphics[width=.99\linewidth]{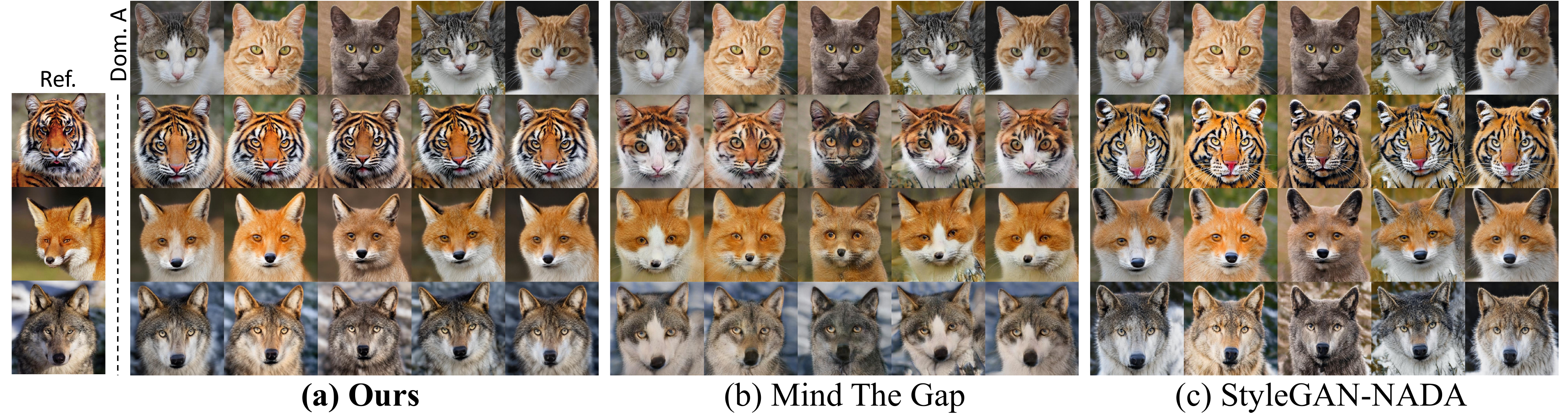}
   \end{center}
  \vspace{-2mm}
   \caption{
   \textbf{Qualitative comparisons using the generator pre-trained on AFHQ-Cat~\cite{choi2020stargan}}.
   The first row shows source images in domain $A$ while the first column presents reference images in domain $B$.
   Our DiFa captures both the representative attributes and styles from various categories of reference images, and exhibits better performance in comparison to the competing methods.
   In contrast, StyleGAN-NADA~\cite{gal2021stylegannada} misses some domain-specific styles while Mind The Gap~\cite{zhu2021mind} fails to obtain the essential attributes of animals in domain $B$.
   \textbf{Results best seen at 500\% zoom.}
   }
    \label{fig:main_cat}
    \vspace{-5mm}
\end{figure}

In this section, we first introduce the experimental settings of our DiFa, including implementation details, datasets, and metrics (\sect{settings}). Both qualitative and quantitative experiments are conducted on a wide range of domains to demonstrate the superiority of our DiFa in generating diverse images and faithful adaption (\sect{results} and \sect{user}). Besides, ablation studies are considered to evaluate the effects of our proposed two losses (\sect{ablation}). Finally, we also investigate the editing ability of the adapted generator and extend our DiFa to zero-shot generative domain adaption (\sect{extension}).

% Firstly, we elaborate the implementation details, datasets and metrics in \sect{settings}.
% Secondly, we demonstrate the superiority of our DiFa by comparing against existing alternatives on qualitative and quantitative experiments (\sect{results}, and a user preference study (\sect{user}), across a wide range of source and target domains.
% Thirdly, we conduct comprehensive ablation studies in \sect{ablation} to verify the effectiveness of our proposed two losses, and more ablation studies about hyper-parameters are provided in the supplementary materials.
% Finally, we investigate the editing capability of the adapted generator and extend our DiFa to zero-shot generative domain adaption in \sect{extension}.
% \vspace{-0.2em}
\subsection{Experimental Settings}
\label{settings}
\myparagraph{Implementation Details.}
In our experiments, we use StyleGAN2 pre-trained on FFHQ~\cite{karras2019style} and StyleGAN-ADA~\cite{karras2020training} pre-trained on AFHQ-Cat~\cite{choi2020stargan}, and employ e4e~\cite{tov2021designing} and pSp~\cite{richardson2021encoding} as their inversion models, respectively.
Following StyleGAN-NADA~\cite{gal2021stylegannada}, we utilize both ViT-B/16 and ViT-B/32~\cite{dosovitskiy2020image} models for CLIP-base losses.
For training, we use ADAM optimizer~\cite{kingma2015adam} with a learning rate 0.02 and set the batch size to 2. 
We finetune the generator for 300$\sim$400 iterations, which takes about 3$\sim$4 minutes on an RTX 2080Ti GPU. 

% In all experiments, we use an ADAM optimizer~\cite{kingma2015adam} with learning rate 0.02 and fine-tune a pre-trained StyleGAN~\cite{karras2020analyzing,karras2020training} with a batch size of two.
% Akin to StyleGAN-NADA~\cite{gal2021stylegannada}, we utilize both ViT-B/16 and ViT-B/32~\cite{dosovitskiy2020image} during training.
% For FFHQ and Cat adaption, we employ e4e~\cite{tov2021designing} and pSp~\cite{richardson2021encoding} as their inversion model respectively.
% Empirically, we fine-tune a generator for 300$\sim$400 iterations, which takes about 3$\sim$4 minutes on a RTX 2080ti.

\myparagraph{Datasets.}
For FFHQ adaption, the target images are collected from three datasets: (i) Artstation-Artistic-face-HQ (AAHQ)~\cite{liu2021blendgan}, (ii) MetFaces~\cite{karras2020training}, and (iii) face paintings by Amedeo Modigliani, Fernand Leger and Raphael~\cite{yaniv2019face}. 
Each of them contains 10 images.
For Cat adaption, we collect target images from the AFHQ-Wild validation dataset and divide them into Tiger, Fox, and Wolf datasets, which include 103, 53, and 46 images, respectively.
In particular, Amedeo Modigliani, Fernand Leger, Raphael, Tiger, Fox and Wolf are used in quantitative experiments.

% We use StyleGAN2 pre-trained on FFHQ~\cite{karras2019style} and StyleGAN-ada~\cite{karras2020training} pre-trained on AFHQ-Cat~\cite{choi2020stargan}.
% The target domains starting from FFHQ mainly come from three datasets: (i) Artstation-Artistic-face-HQ (AAHQ)~\cite{liu2021blendgan}, (ii) MetFaces~\cite{karras2020training}, and (iii) face paintings by Amedeo Modigliani, Fernand Leger and Raphael~\cite{yaniv2019face}, each of which only contains ten images.
% For Cat domain adaption, we collect images from AFHQ-Wild validation dataset and divide them into Tiger, Fox and Wolf datasets, which include 103, 53 and 46 images, respectively.
% In particular, Amedeo Modigliani, Fernand Leger, Raphael, Tiger, Fox and Wolf are used in quantitative experiments.

\myparagraph{Metrics.}
Following StyleGAN-ADA~\cite{karras2020training}, we use \fid (FID)~\cite{heusel2017gans} and Kernel Inception Distance (KID)~\cite{binkowski2018demystifying} to evaluate our DiFa quantitatively. 
Both metrics measure the quality and diversity of the images, while KID is more suitable for the few-shot setting (only a few images in validation sets).
In all our experiments, both FID and KID are calculated between 5,000 synthesized images and each validation sets.

%
% FID measures the similarity of two sets in the high-dimensional feature space given by an InceptionV3~\cite{simonyan2014very}.
%
% KID computes the squared maximum mean discrepancy to measure the visual similarity of two sets.
%
% According to StyleGAN-ADA~\cite{karras2020training}, \fid (FID) does not serve as an ideal metric for small datasets.
% %
% When there are only a few images in validation sets (\eg, ten samples in face paintings), FID is dominated by the inherent bias and the generator which simply copies training images will be seen as a perfect model.
% %
% Therefore, following StyleGAN-ada~\cite{karras2020training}, we employ Kernel Inception Distance (KID)~\cite{binkowski2018demystifying} as the main metric while reporting FID scores, and they both are calculated between 5,000 generated samples and all validation sets.
% \vspace{-0.4em}
\subsection{Qualitative and Quantitative Evaluation}
\label{results}
\myparagraph{Qualitative Results.}

\fig{fig:main_ffhq} shows the qualitative comparisons adapted from FFHQ~\cite{karras2019style}.
As shown in the figure, Few-Shot Adaption~\cite{ojha2021few} suffers from severe model collapse and generates similar images.
Due to StyleGAN-NADA~\cite{gal2021stylegannada} is trained by aligning the sample-shift direction $\Delta \vv_{samp}$ with domain-gap direction $\Delta \vv_{dom}$, which contains the domain-sharing attributes (\eg, gender) shift, it also cannot inherit the sufficient diversity from the pre-trained generator. 
For example, the gender of adapted images is changed to female in 4$\sim$6-th rows in \fig{fig:main_ffhq}(c).
%
%
% StyleGAN-NADA~\cite{gal2021stylegannada} cannot retain the common domain-sharing attributes, \eg, gender and hair length in 4th row of \fig{fig:main_ffhq}(c), generating images with insufficient diversity.
% cannot acquire sufficient domain-specific styles, \eg, white color of sculptures in first row of \fig{fig:main_ffhq}(c).
Mind The Gap~\cite{zhu2021mind} retains the local styles of reference image via style mixing. 
However, it produces undesired semantic artifacts when there is a significant shape discrepancy between domains, \eg, redundant noses and eyes in 3rd and last row of \fig{fig:main_ffhq}(b). 
In contrast, with the proposed SCC and AS losses, our DiFa not only faithfully acquires the representative domain-specific attributes and styles from the reference image, but also produces images with high diversity inherited from the pre-trained generator.
Additionally, we also illustrate the qualitative results adapted from AFHQ-Cat~\cite{choi2020stargan} in \fig{fig:main_cat}.
Our DiFa also captures sufficient domain-specific characteristics from the reference image in comparison to the competing methods (\eg, the mane and stripes of the tiger in first row), further demonstrating the superiority of our method.
More visualizations adapted from other domains are shown in \emph{Suppl}.

% %
% \fig{fig:main_ffhq} and \fig{fig:main_cat} illustrate the qualitative comparisons with existing state-of-the-art methods~\cite{zhu2021mind, gal2021stylegannada, ojha2021few}, whose source domains are FFHQ~\cite{karras2019style} and AFHQ-Cat~\cite{choi2020stargan}, respectively.
% %
% As shown in \fig{fig:main_ffhq}, our DiFa generates images with higher diversity and less visible artifacts, surpassing previous approaches in a large margin.
% %
% On the one hand, StyleGAN-NADA~\cite{gal2021stylegannada} and Few-Shot Adaption~\cite{ojha2021few} cannot retain the domain-sharing attributes like hairstyles, and thus there is insufficient diversity for both of them.
% %
% Especially, Few-Shot Adaption~\cite{ojha2021few} suffers from severe mode collapse and performs the worst.
% %
% On the other hand, Mind The Gap~\cite{zhu2021mind} fails to acquire some domain-specific attributes, leading to unfaithful attribute-level adaption.
% %
% Compared with other alternatives, \fig{fig:main_ffhq} and \fig{fig:main_cat} indicate that our DiFa also captures more sufficient domain-specific styles from the reference image.
% %
% Take the first row of \fig{fig:main_cat} as an example, images from our DiFa have the similar mane and stripes with the tiger in reference image, whereas the other two do not.
% %
% As a contrast, Mind The Gap~\cite{zhu2021mind} synthesises images with undesired semantic artifacts when there is a significant shape change, \eg, the third and last row of \fig{fig:main_ffhq}(b).

\myparagraph{Quantitative Results.}
\begin{table}[t]
% \vspace{-1em}
\caption{
\textbf{KID ($\downarrow$) comparisons between different one-shot domain adaption methods.}
Each result is averaged over 5 training shots and in the form of $\{$mean $\pm$ standard error$\}$.
FSA, NADA and MTG denote Few-Shot Adaption~\cite{ojha2021few}, StyleGAN-NADA~\cite{gal2021stylegannada} and Mind The Gap~\cite{zhu2021mind}, respectively.}
\vspace{2mm}
\centering
\scalebox{0.8}{
\begin{tabular}{l|ccc|ccc}
\toprule
\multicolumn{1}{l}{\multirow{2}{*}{Models}} 
& \multicolumn{3}{c}{FFHQ}
& \multicolumn{3}{c}{Cat}\\
\cmidrule(l){2-4}
\cmidrule(l){5-7}
{} & {Amedeo.} & {Fernand.} & {Raphael} & {Tiger} & {Fox} & {Wolf} \\
\midrule
FSA~\cite{ojha2021few} &180.10$\pm$ 1.12  &187.26$\pm$ 13.10  &165.25$\pm$ 66.31  & -  & -  & - \\
% \midrule
NADA~\cite{gal2021stylegannada} &131.03$\pm$ 28.14 &169.83$\pm$ 31.52 &149.19$\pm$ 55.91 &13.83$\pm$ 2.75 &73.17$\pm$ 39.30 &47.96$\pm$ 20.37 \\
% \midrule
MTG~\cite{zhu2021mind} &146.84$\pm$ 46.24 &192.19$\pm$ 42.73 &125.58$\pm$ 18.63 &48.27$\pm$ 13.87 &69.19$\pm$ 26.23 &51.11$\pm$ 11.32\\
% \midrule
\midrule
\textbf{Ours} &\textbf{121.21 $\pm$ 24.62} &\textbf{159.93 $\pm$ 31.39} &\textbf{112.72 $\pm$ 17.61} &\textbf{13.13 $\pm$ 2.09} &\textbf{54.20 $\pm$ 31.42} &\textbf{33.52 $\pm$ 9.21} \\
\bottomrule
\end{tabular}
}
\label{tab:main_kid}
% \vspace{-4mm}
\end{table}
\begin{table}[t]
\caption{
\textbf{FID ($\downarrow$) comparisons between different one-shot domain adaption methods.}
Each result is averaged over 5 training shots and in the form of $\{$mean $\pm$ standard error$\}$.}
\vspace{2mm}
\centering
\scalebox{0.8}{
\begin{tabular}{l|ccc|ccc}
\toprule
\multicolumn{1}{l}{\multirow{2}{*}{Models}} 
& \multicolumn{3}{c}{FFHQ}
& \multicolumn{3}{c}{Cat}\\
\cmidrule(l){2-4}
\cmidrule(l){5-7}
{} & {Amedeo.} & {Fernand.} & {Raphael} & {Tiger} & {Fox} & {Wolf} \\
\midrule
FSA~\cite{ojha2021few} &\textbf{171.56} $\pm$ \textbf{33.68}  &\textbf{236.61}$\pm$ \textbf{25.03}  &177.47$\pm$ 32.21  & -  & -  & - \\
% \midrule
NADA~\cite{gal2021stylegannada} &188.44$\pm$ 19.15 &257.27$\pm$ 19.39 &186.20$\pm$ 28.60 &16.74 $\pm$ 1.53 &82.59 $\pm$ 25.31 &54.28$\pm$ 13.34 \\
% \midrule
MTG~\cite{zhu2021mind} &215.88$\pm$ 34.14 &278.46$\pm$ 48.27 &193.76$\pm$ 7.07 &46.72$\pm$ 13.34 &82.30$\pm$ 15.09 &58.65$\pm$ 6.60\\
% \midrule
\midrule
\textbf{Ours} &187.28 $\pm$ 24.45 &254.68 $\pm$ 17.73  &\textbf{172.34 $\pm$ 10.15} &\textbf{16.26 $\pm$ 1.08} &\textbf{71.57 $\pm$ 18.18} &\textbf{44.39 $\pm$ 5.96} \\
\bottomrule
\end{tabular}
}
\label{tab:app_fid}
% \vspace{-5.5mm}
\end{table}

We also quantitatively compare our DiFa with competing methods~\cite{ojha2021few, zhu2021mind, gal2021stylegannada} under six settings, \ie, $\hbox{FFHQ} \rightarrow \hbox{\{Amedeo Modigliani, Fernand Leger, Raphael\}}$ and $\hbox{Cat} \rightarrow \hbox{\{Tiger, Fox, Wolf\}}$.
For each setting, we randomly sample an image from a target dataset to perform adaption, and report both Kernel Inception Distance (KID)~\cite{binkowski2018demystifying} and \fid (FID)~\cite{heusel2017gans}  metrics.
To reduce random sampling error, we repeat it five times and use the mean value as final score.
The results are listed in Table~\ref{tab:main_kid} and Table~\ref{tab:app_fid}. 
One can see that our DiFa clearly outperforms the competing methods, which are consistent with qualitative results in \fig{fig:main_ffhq} and \fig{fig:main_cat}.
We observe that FSA~\cite{ojha2021few} obtains better FID scores in Amedeo and Fernand datasets, which is inconsistent with above qualitative results (see \fig{fig:main_ffhq}(d)).
Note that FID cannot reflect the overfitting problem very well when target dataset is extremely small and biased~\cite{karras2020training}.
Specifically, these two small datasets have different data biases with FFHQ, \eg, gender bias.
8/10 images in the Amedeo Modigliani dataset and 9/10 images in the Fernand Leger dataset are female. 
Due to our DiFa acquiring the diversity from the original generator which is trained on FFHQ, it generates male and female adapted images with similar probability. 
In contrast, for FSA, the adapted images are all similar to the reference image. 
When comparing on the above two datasets, FSA tends to generate images that have similar gender distribution to the validation dataset, thus achieving better FID results. 
For the Raphael dataset, which has 5/10 images that are female, our DiFa achieves better FID results.
%

% We evaluate different methods on one-shot domain adaption under six settings, \ie, $\hbox{FFHQ} \rightarrow \hbox{\{Amedeo Modigliani, Fernand Leger, Raphael\}}$ and $\hbox{Cat} \rightarrow \hbox{\{Tiger, Fox, Wolf\}}$.
% For each setting, we randomly sample an image from a target dataset to perform adaption, and repeat it five times to reduce the error of random sampling.
% In Table~\ref{tab:main_kid}, experimental results illustrate that our framework clearly outperforms all existing approaches from the perspective of KID, which are consistent with qualitative results shown in \fig{fig:main_ffhq} and \fig{fig:main_cat}.
% Albeit FID cannot reflect overfitting problem very well~\cite{karras2020training} when target dataset is extremely small, we still report it in Table~\ref{tab:app_fid}.
% We observe that our DiFa remains to achieve better performance than multiple alternatives.
% It is noted that FSA~\cite{ojha2021few} obtains the lowest FID scores in Amedeo and Fernand datasets, however, due to its severe mode collapse (\fig{fig:main_ffhq})

% Albeit FID scores cannot reflect overfitting problem very well, we present FID scores in Table~\ref{tab:app_fid}, and observe that our approach remains to achieve better performance than multiple alternatives.
% Since FSA~\cite{ojha2021few} produces images almost same as the reference (See \fig{fig:main_ffhq}(d)), it obtains lower FID scores when a target dataset is extremely small, \eg, Amedeo Modigliani with ten images only.

% \vspace{-0.5em}
\subsection{User Study}
% \vspace{-0.3em}
\label{user}
\begin{table}[t]
\vspace{-1em}
\caption{
    \textbf{User preference study.}
    The numbers represent the percentage of users who favor the images synthesized by our DiFa over the other competitor.}
    \vspace{1mm}
    \centering
    \scalebox{0.9}{
    \begin{tabular}{l|c|c|c}
         \makecell{Model\\Comparison} & \makecell{Image \\ Quality} & \makecell{Style\\Similarity} & \makecell{Attribute\\Consistency} \\
         \Xhline{\arrayrulewidth}
         Ours vs. FSA~\cite{ojha2021few} & 87.90$\%$ & 36.00$\%$ & 96.10$\%$ \\
         Ours vs. NADA~\cite{gal2021stylegannada} & 76.76$\%$ & 77.33$\%$ &78.95$\%$\\
         Ours vs. MTG~\cite{zhu2021mind} & 81.43$\%$ & 73.62$\%$ & 64.05$\%$ \\
    \end{tabular}
    }
    \vspace{-2mm}
    \label{tab:main_user}
\end{table}

We further perform user study to compare our DiFa with the competing methods.
Specifically, we provide users a reference image, a source image, and two adapted images from different methods, and ask them to choose the better adapted image for each of three measurements: (i) image quality, (ii) style similarity with the reference and (iii) attribute consistency with the source image.
We randomly generate 1,050 samples for each comparison (3,150 in total).
There are 30 users. We assign 105 samples for each of them, and give them unlimited time to complete the evaluation.
From Table~\ref{tab:main_user}, the users strongly favor our DiFa in all three aspects, especially from the perspective of image quality and attribute consistency.
Note that FSA~\cite{ojha2021few} suffers from severe mode collapse and simply copies from the reference, hence, it is favored on style similarity but performs worse on the other aspects.

% \vspace{-0.3em}
\subsection{Ablation Study}
% \vspace{-0.3em}
\label{ablation}
\begin{figure}[t]
   % \vspace{-1em}
   \begin{center}
   \includegraphics[width=.99\linewidth]{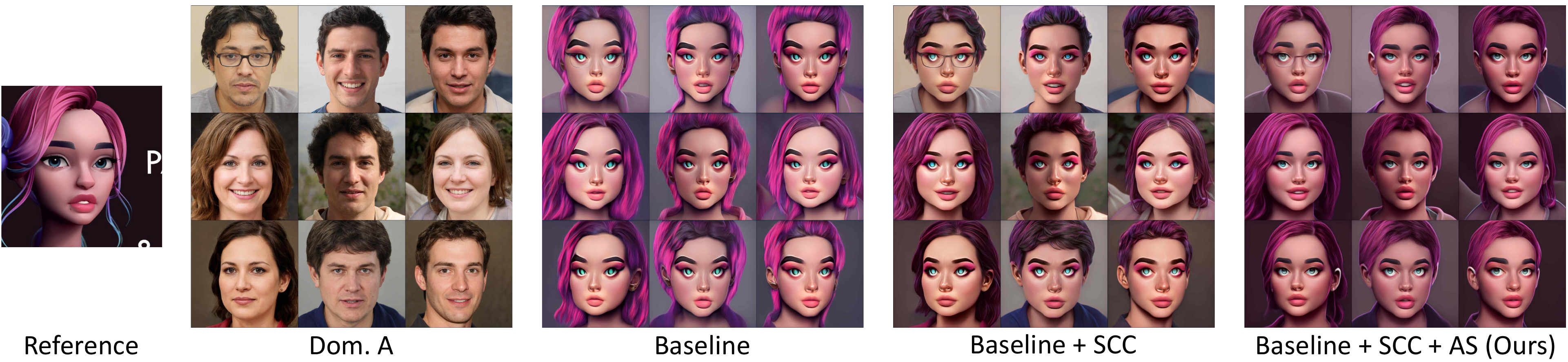}
   % \fbox{\rule{0pt}{2in} \rule{.9\linewidth}{0pt}} 
   \end{center}
  \vspace{-2mm}
   \caption{
   \textbf{Ablation studies on Selective Cross-modal Consistency (SCC) loss and Attentive Style (AS) loss}.
   Compared to the baseline, SCC enhances keeping the consistency (\eg, hairstyles) between source and adapted images, vastly boosting the diversity of generation.
   Moreover, AS encourages the adapted generator to further acquire the representative styles (\eg, darker hue and purple hair) from the reference.}
    \label{fig:main_ablation}
    \vspace{-4mm}
\end{figure}
Ablation studies are conducted to evaluate the effects of two critical components of our DiFa, \ie, the selective cross-domain consistency (SCC) loss and the attentive style (AS) loss.
As shown in \fig{fig:main_ablation}, the images from the baseline have very limited diversity and lack some representative characteristics of the reference image, \eg, darker hue.
% As shown in \fig{fig:main_ablation}, our baseline only obtains the global-level attributes (\eg, big eyes and thick lips) from the reference, however, failing to produce images with both high diversity and the representative styles of the reference image.
Benefited from SCC, the adapted generator begins to retrain the domain-sharing attributes (\eg, hair length and gender), thereby inheriting the diverse generation ability from the pre-trained generator.
When further adding AS, we observe that the adapted generator faithfully captures the domain-specific styles and local-level representative attributes from the reference image, \eg, darker hue and purple hair. 
More ablation studies about hyper-parameters are provided in the \emph{Suppl}.

% Compared to the baseline, SCC encourages the adapted generator to retain more domain-sharing attributes (\eg, hairstyles), so that the adapted generator inherits the capability in producing more diverse images from the pre-trained generator while acquiring domain-specific attributes.
% When further adding AS during adaption, we observe that adapted images capture more faithfully domain-specific styles from the reference, \eg, darker hue and purple hair. 
% More ablation studies about hyper-parameters are provided in the \emph{Suppl}.
% \vspace{-0.3em}
\subsection{Extensions}
\label{extension}
\begin{figure}[t]
% \vspace{-1em}
   \begin{center}
   \includegraphics[width=.9\linewidth]{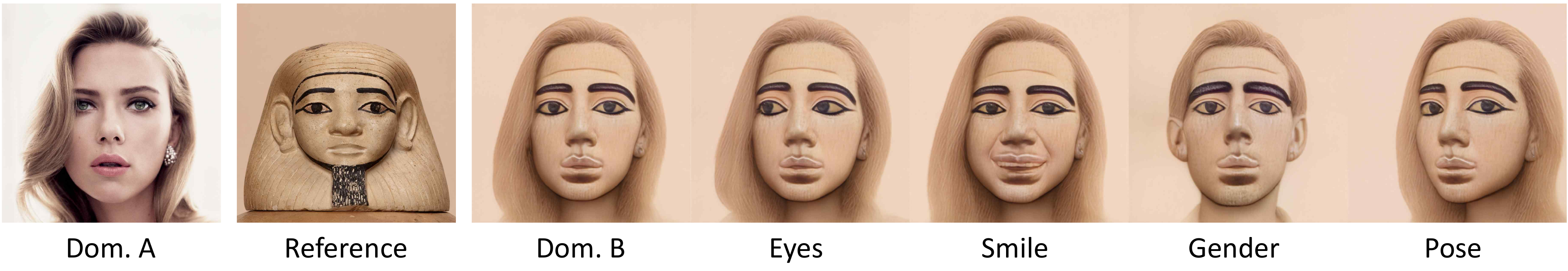}
   \end{center}
   \vspace{-2mm}
   \caption{
   \textbf{Editing a real image in domain $B$.}
   The first three columns show a real image in domain $A$, a reference image, and an adapted real image in domain $B$, respectively.
   The other columns present the editing operations and their corresponding results in domain $B$.
   All editing directions are discovered by InterfaceGAN~\cite{shen2020interfacegan} in domain $A$.
   }
    \label{fig:img_edit}
    \vspace{-2mm}
\end{figure}
\begin{figure}[t]
   \begin{center}
   \includegraphics[width=.99\linewidth]{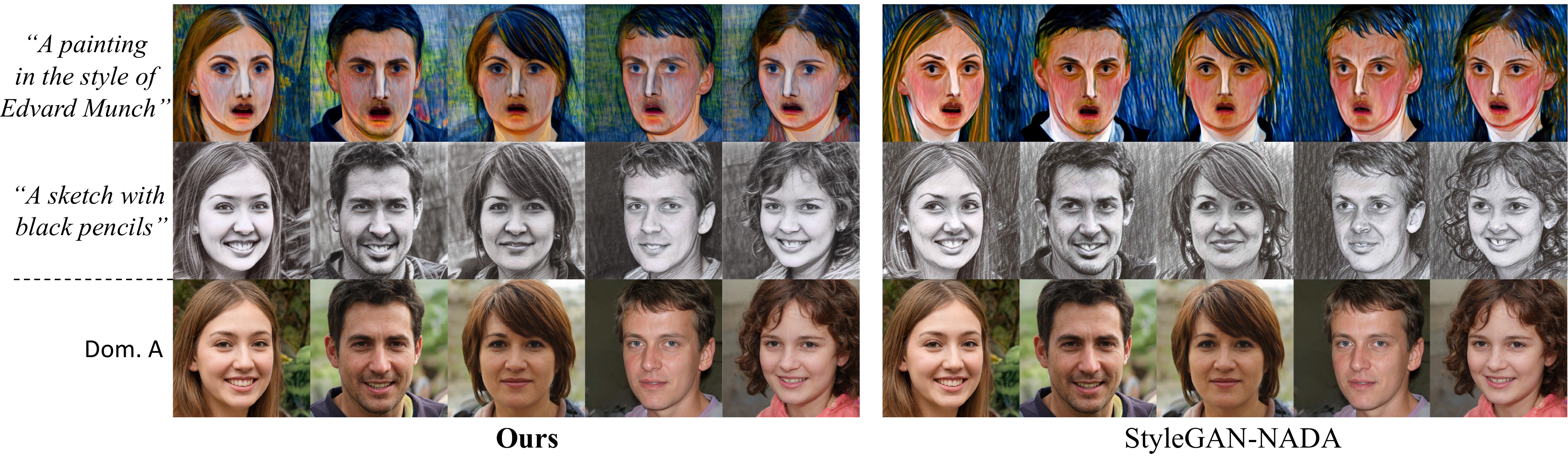}
   \end{center}
  \vspace{-2mm}
   \caption{
   \textbf{Qualitative comparisons on zero-shot generative domain adaption} between our DiFa and StyleGAN-NADA~\cite{gal2021stylegannada}.
   Given the text description in first column, our DiFa adapts source images in domain $A$ (last row) to the described target domain, significantly surpassing StyleGAN-NADA~\cite{gal2021stylegannada} from the consistency perspective.
    }
    \label{fig:zero_shot}
    \vspace{-1em}
\end{figure}
\myparagraph{Latent Space Editing.}
In \fig{fig:img_edit}, we illustrate the editing results performed on a real image adapted into a new domain.
Concretely, we employ InterfaceGAN~\cite{shen2020interfacegan} to discover some editing directions in domain $A$, and then leverage these directions to edit the adapted real image.
As can be seen, the directions from domain $A$ still manage to control real images in domain $B$, indicating that the adapted generator maintains a similar ability in latent-based editing with the original generator.

% We note that editing only involves in first eight layers of $\gW+$ to prevent the domain-specific styles from being disrupted.

% To investigate the editing capability in the adapted latent space, we intend to adapt and edit a real image  domain B and then edit it with existing image editing approaches~\cite{shen2020interfacegan}.
% Concretely, we leverage InterfaceGAN~\cite{shen2020interfacegan} to discover some editing directions in domain A, including eyes, smile, gender and pose.
% Next, we employ e4e~\cite{tov2021designing} to invert a real image into $\gW+$ code, and only edit the first eight layers of this code to prevent domain-specific styles from being disrupted.
% \fig{fig:img_edit} shows the editing results of a real image in domain B.
% As can be seen, the directions from domain A still manage to control real images in domain B, indicating that the adapted generator well inherits the editing latent space from the original generator.
% and there is not need to re-discover such directions when adapting to a new domain.

% \vspace{-1mm}
\myparagraph{Zero-shot Domain Adaption of GANs.}
With minor modifications (\eg, removing the AS loss), our DiFa can be easily extended to zero-shot GDA, \ie, adapting to a target domain described by text only.
\fig{fig:zero_shot} shows the comparison between our DiFa and StyleGAN-NADA~\cite{gal2021stylegannada}.
One can see that adapted images from our DiFa are more consistent with their corresponding source images, thereby inheriting more diversity from the pre-trained generator.
For example, when performing adaption from FFHQ to a target domain described by ``\textit{A sketch with black pencils}'', all eyes in StyleGAN-NADA results look to the left, which is inconsistent with the original eyes in source domain.
More visualizations and the implementation details are given in the \emph{Suppl}.
% \vspace{-0.7em}
\section{Discussion}
% \vspace{-0.7em}
\label{discussion}
In this paper, we presented DiFa to address the diverse generation and faithful adaptation issues for one-shot generative domain adaption.
In particular, DiFa leverages the difference between the CLIP embedding of the reference image and the embedding of source domain to guide the global-level adaption.
To faithfully acquire local-level domain-specific characteristics, we introduce the attentive style loss to align each intermediate token of adapted images with its closest token of the reference image.
For highly diverse generation, the selective cross-domain consistency loss is proposed to select and retain the domain-sharing attributes in $\gW+$ space.
Both qualitative and quantitative experiments show the superiority of our DiFa against state-of-the-arts under a wide range of settings, especially for the cases of large domain gap.
Furthermore, our DiFa can be easily extended to zero-shot generative domain adaption with compelling results.

% \vspace{-1mm}
\myparagraph{Limitations.}When there are few domain-sharing attributes between source and target domains, \eg, $\hbox{Church} \rightarrow \hbox{Tiger}$, our DiFa cannot produce highly diverse images.
Fortunately, this issue may be largely alleviated by adaptively inheriting the prior knowledge from large-scale generators~\cite{Sauer2021stylegan_xl,ramesh2021zero,nichol2021glide}, which are pre-trained on sufficient variety of source domains.
% Despite advanced performance on one-shot generative domain adaption, our DiFa sometimes fails to deal with the cases where there are few domain-sharing attributes.
% For example, when adapting a generator pre-trained in church domain to tiger domain, the adapted generator may have little prior knowledge to leverage and thus produces images with very low diversity.
% Fortunately, using the generator pre-trained on a wide range of source domains~\cite{Sauer2021stylegan_xl}, we adaptively select the source domain closest to the reference, so that the adapted generator could inherit more diversity from the pre-trained generator.

% \vspace{-1mm}
\myparagraph{Broader Impact.}
Transferring a pre-trained generator with very limited data plays a crucial role in academia and industry.
More specifically, our DiFa provides insights on tasks in computer vision, \eg, data augmentation and few-shot adaption.
Meanwhile, our DiFa also makes AI more accessible to the public.
On the one hand, users could leverage our method to create the artworks with any desired styles, even without adequate computing and data resources.
On the other hand, our work may bring potential concerns on the probability of producing fake images.
For example, someone may use our DiFa to spoof other people's portraits, to synthesize deceptive interactions, or even to impersonate public figures to influence political processes.
Albeit there are a few potential negative impacts, we believe that they could be well addressed with the development of DeepFake detection and proper protocols.
In particular, we could verify the authenticity, integrality, and source of images by adding digital watermarks or signatures.
Also, we may employ DeepFake detection technique to analyze the images without digital signatures.
Furthermore, our community should help the government to improve corresponding laws and regulations to avoid the abuse of image generation.

% Transferring a pre-trained generator to a new domain with one reference image is particularly beneficial for both practical applications and the development of the community.
% Firstly, it provides insights on how to effectively inherit the prior knowledge from large scale pre-trained models, like DALL-E~\cite{ramesh2021zero} and GLIDE~\cite{nichol2021glide}.
% Secondly, it introduces a novel approach of data augmentation.
% There are many research fields where there is a lack of training samples, and our approach could expand the size and diversity of training data significantly.
% Thirdly, it promotes the creation of artworks and makes AI more accessible to public.
% With our DiFa, users could transfer real images to any desired style with less time and lower computing resources.
% However, our work may brings potential concerns on the ability to produce fake images, but we believe that these concerns could be addressed with the development of DeepFake detection.

% \vspace{-3mm}\paragraph{}
% \newpage
{\small
\bibliographystyle{ieee_fullname}
\bibliography{bib}
}

\section*{Checklist}
\begin{enumerate}

\item For all authors...
\begin{enumerate}
  \item Do the main claims made in the abstract and introduction accurately reflect the paper's contributions and scope?
    \answerYes{}  The main contributions can be summarized as: (1) We introduce a novel method namely DiFa, along with Selective Cross-domain Consistency Loss (SCC) and Attentive Style Loss (AS), for diverse generation and faithful adaption.
    (2) Extensive experiments highlight the effectiveness of our DiFa in acquiring the representative characteristics from the reference image, and inheriting the capability to produce high-diversity images from the pre-trained generator.
    (3) Our DiFa can be easily extended to zero-shot generative domain adaption with appealing results.
  \item Did you describe the limitations of your work?
    \answerYes{} See \sect{discussion}
  \item Did you discuss any potential negative societal impacts of your work?
    \answerYes{} See \sect{discussion}
  \item Have you read the ethics review guidelines and ensured that your paper conforms to them?
    \answerYes{} We carefully read the ethics review and comply with them.
\end{enumerate}

\item If you are including theoretical results...
\begin{enumerate}
  \item Did you state the full set of assumptions of all theoretical results?
    \answerNA{}
        \item Did you include complete proofs of all theoretical results?
    \answerNA{}
\end{enumerate}

\item If you ran experiments...
\begin{enumerate}
  \item Did you include the code, data, and instructions needed to reproduce the main experimental results (either in the supplemental material or as a URL)?
    \answerYes{} See \url{https://github.com/1170300521/DiFa}.
  \item Did you specify all the training details (e.g., data splits, hyper-parameters, how they were chosen)?
    \answerYes{} See Sec.~\ref{overall}, Sec.~\ref{settings} and \url{https://github.com/1170300521/DiFa}
        \item Did you report error bars (e.g., with respect to the random seed after running experiments multiple times)?
    \answerYes{} See Table~\ref{tab:main_kid} and Table~\ref{tab:app_fid}
        \item Did you include the total amount of compute and the type of resources used (e.g., type of GPUs, internal cluster, or cloud provider)?
    \answerYes{} See Sec.~\ref{experiments}
\end{enumerate}

\item If you are using existing assets (e.g., code, data, models) or curating/releasing new assets...
\begin{enumerate}
  \item If your work uses existing assets, did you cite the creators?
    \answerYes{} We cite papers that we used in Sec.~\ref{experiments}
  \item Did you mention the license of the assets?
    \answerYes{} We use AFHQ, MetFaces, and Artistic-Faces datasets during training, and their license are included in our repository \url{https://github.com/1170300521/DiFa}. 
  \item Did you include any new assets either in the supplemental material or as a URL?
    \answerYes{} We provide the URL of our source code in the supplemental material.
  \item Did you discuss whether and how consent was obtained from people whose data you're using/curating?
    \answerYes{} The AFHQ, MetFaces, and Artistic-Faces datasets are widely used in the community. To our best knowledge, there is no inappropriate information in them.
  \item Did you discuss whether the data you are using/curating contains personally identifiable information or offensive content?
    \answerYes{} The datasets we used in this paper are popular benchmarks in computer vision and do not include any harmful information.
\end{enumerate}

\item If you used crowdsourcing or conducted research with human subjects...
\begin{enumerate}
  \item Did you include the full text of instructions given to participants and screenshots, if applicable?
    \answerYes{} See Sec.~\ref{user}
  \item Did you describe any potential participant risks, with links to Institutional Review Board (IRB) approvals, if applicable?
    \answerNA{}
  \item Did you include the estimated hourly wage paid to participants and the total amount spent on participant compensation?
    \answerYes{} Each user is paid 10 dollars per hour and we spend about 300 dollars in total.
\end{enumerate}

\end{enumerate}

\clearpage
\begin{appendices}
\section*{Appendix}
\section*{A \quad Outline}
\vspace{-1em}
Our code is available at \url{https://github.com/1170300521/DiFa}.
In this appendix, we begin to show more visualization results for zero-shot and one-shot generative domain adaption in Sec. \textcolor{red}{B}.
Additionally, Sec. \textcolor{red}{C} presents implementation details under the zero-shot setting.
Furthermore, we conduct more ablation studies and comparison experiments in Sec. \textcolor{red}{D} and Sec. \textcolor{red}{E} respectively.
Finally, we elaborate the user study in Sec. \textcolor{red}{F}.

\vspace{-0.8em}
\section*{B \quad More Visualizations}
\begin{figure}[t]
   \vspace{-1em}
  \begin{center}
  \includegraphics[width=.99\linewidth]{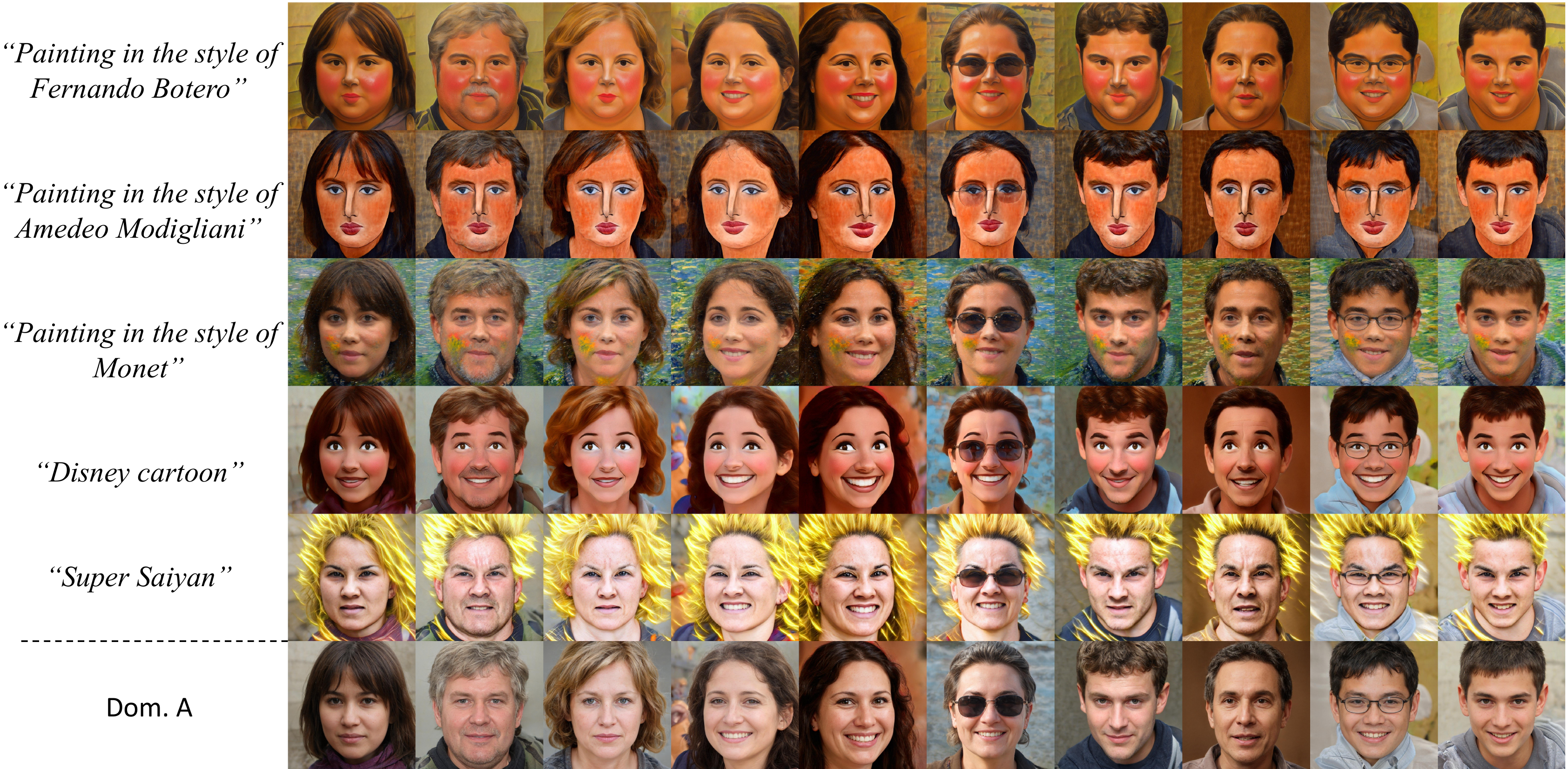}
  \end{center}
  \vspace{-2mm}
  \caption{
  \textbf{Qualitative results for zero-shot generative domain adaption.}
    The last row and first column show the source images from domain $A$ and the descriptions of domain $B$, respectively.
    \textbf{Results best seen at 500\% zoom.}
  }
    \label{fig:supp_zero}
  \vspace{-2.5mm}
\end{figure}

\begin{figure}[t]
   \vspace{-1em}
  \begin{center}
  \includegraphics[width=.99\linewidth]{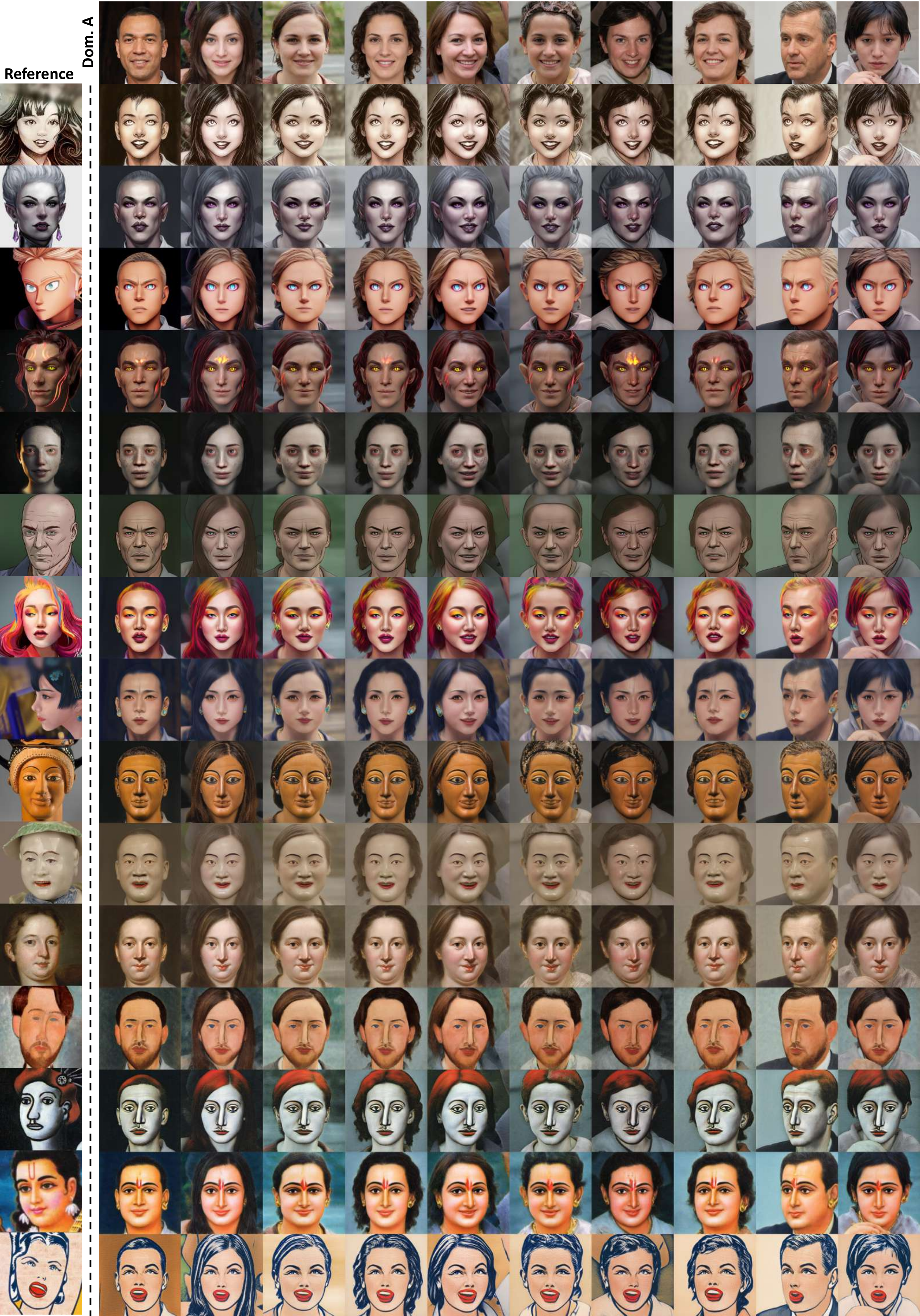}
  \end{center}
  \vspace{-2mm}
  \caption{
  \textbf{Qualitative results using the generator pre-trained on FFHQ.}
    The first row and column show the source images from domain $A$ and the reference images from domain $B$, respectively.
    \textbf{Results best seen at 500\% zoom.}
  }
    \label{fig:supp_ffhq}
  \vspace{-2.5mm}
\end{figure}

\begin{figure}[t]
   \vspace{-1em}
  \begin{center}
  \includegraphics[width=.99\linewidth]{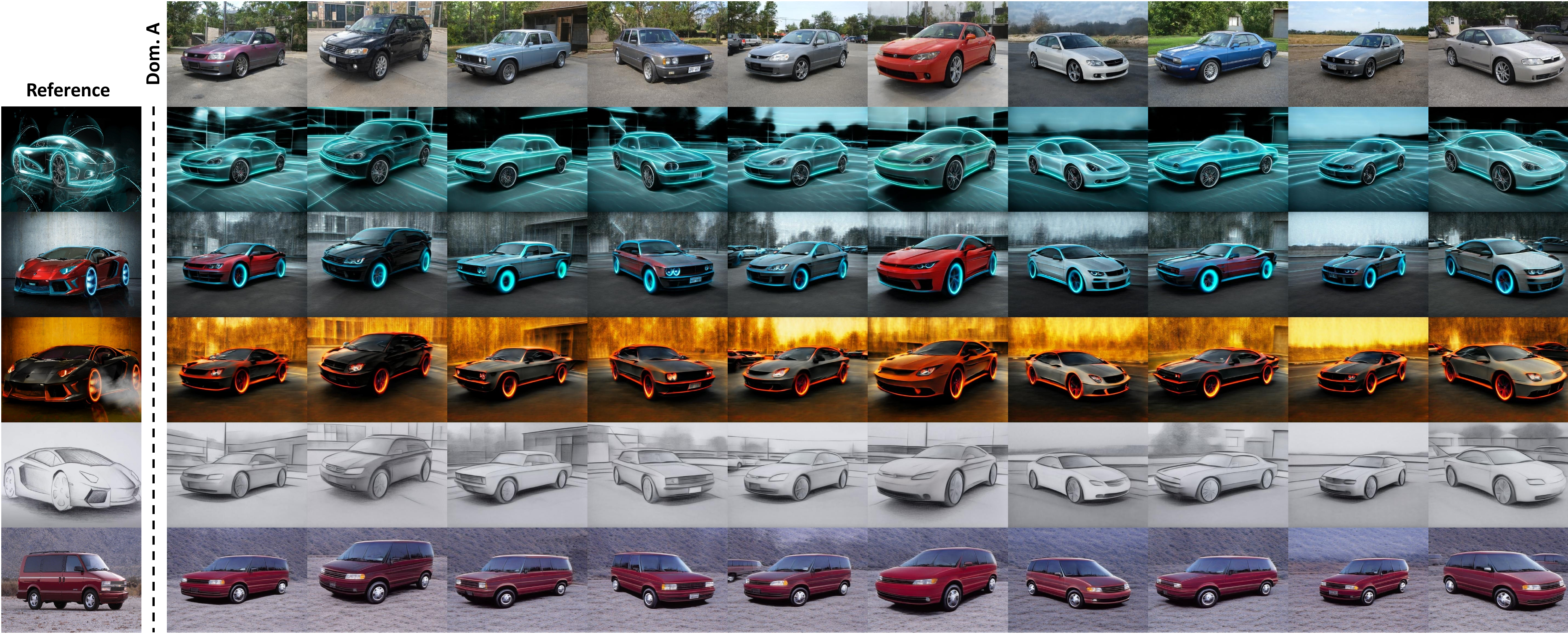}
  \end{center}
  \vspace{-2mm}
  \caption{
  \textbf{Qualitative results using the generator pre-trained on LSUN CAR.}
    The first row and column show the source images from domain $A$ and the reference images from domain $B$, respectively.
    \textbf{Results best seen at 500\% zoom.}
  }
    \label{fig:supp_car}
  \vspace{-2.5mm}
\end{figure}

\begin{figure}[t]
   \vspace{-1em}
  \begin{center}
  \includegraphics[width=.99\linewidth]{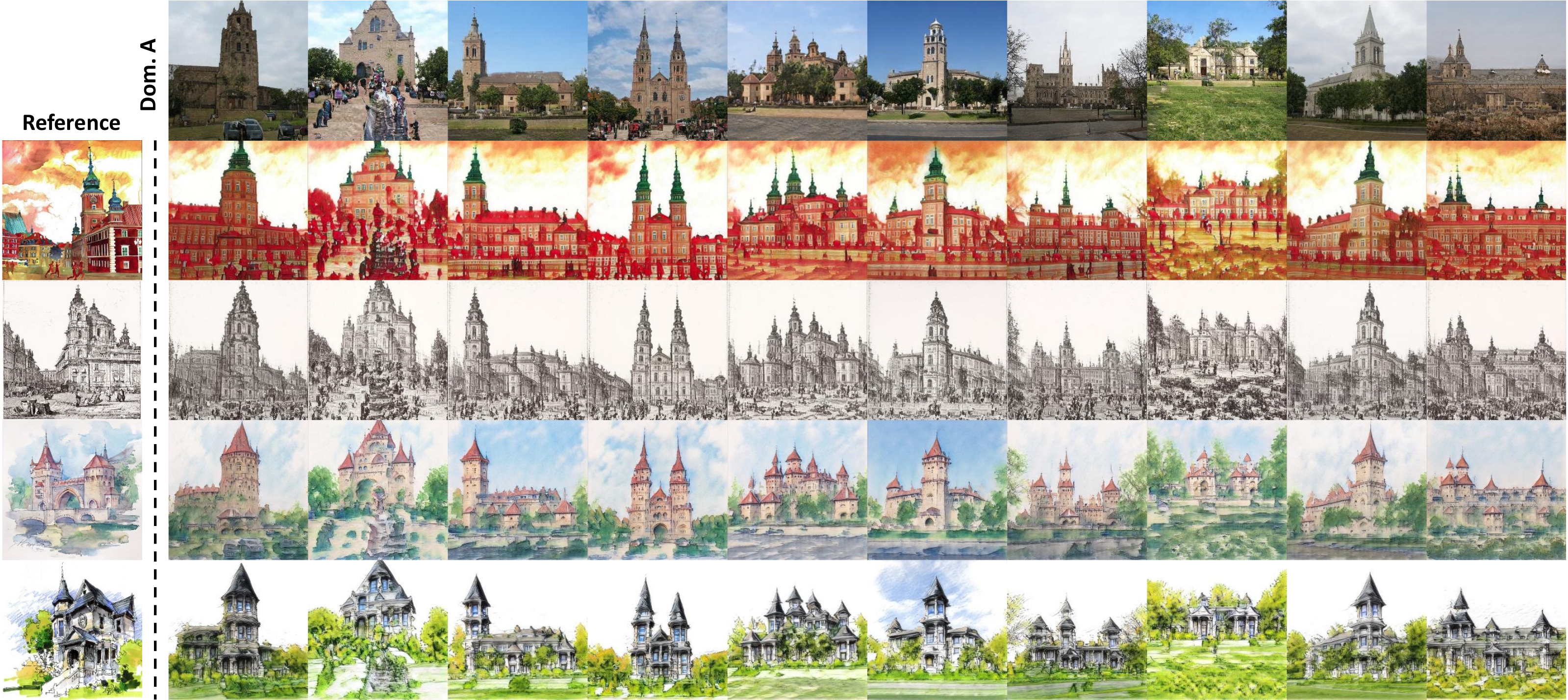}
  \end{center}
  \vspace{-2mm}
  \caption{
  \textbf{Qualitative results using the generator pre-trained on LSUN CHURCH.}
    The first row and column show the source images from domain $A$ and the reference images from domain $B$, respectively.
    \textbf{Results best seen at 500\% zoom.}
  }
    \label{fig:supp_church}
  \vspace{-2.5mm}
\end{figure}

\begin{figure}[t]
   \vspace{-1em}
  \begin{center}
  \includegraphics[width=.99\linewidth]{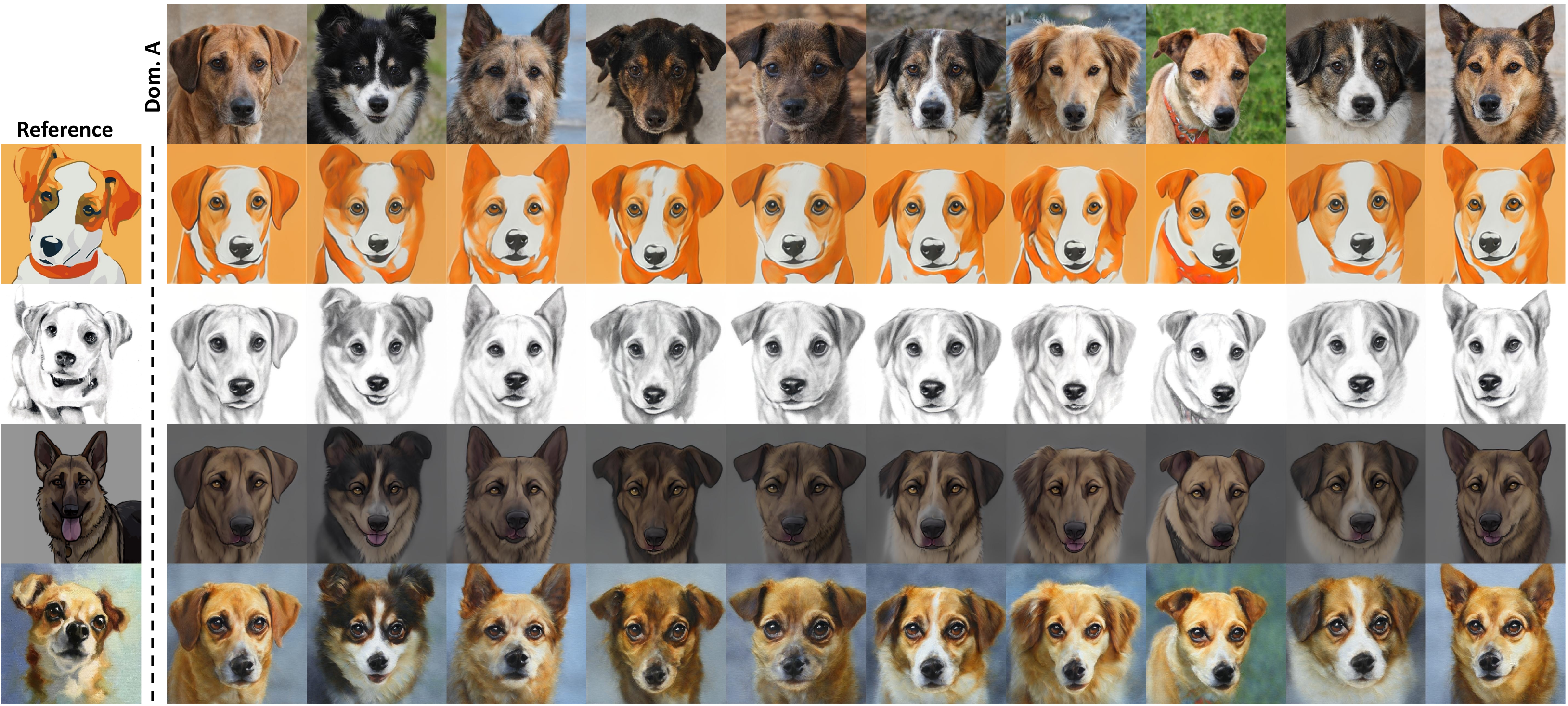}
  \end{center}
  \vspace{-2mm}
  \caption{
  \textbf{Qualitative results using the generator pre-trained on AFHQ-Dog.}
    The first row and column show the source images from domain $A$ and the reference images from domain $B$, respectively.
    \textbf{Results best seen at 500\% zoom.}
  }
    \label{fig:supp_dog}
  \vspace{-2.5mm}
\end{figure}

\vspace{-0.8em}
\label{app_results}
We provide more qualitative results from a wide range of source and target domains.
Fig.~\ref{fig:supp_ffhq} shows the results converted from the generator pre-trained on FFHQ.
Fig.~\ref{fig:supp_car} shows the results converted from the generator pre-trained on LSUN CAR.
Fig.~\ref{fig:supp_church} shows the results converted from the generator pre-trained on LSUN CHURCH.
Fig.~\ref{fig:supp_dog} shows the results converted from the generator pre-trained on AFHQ-Dog.
Fig.~\ref{fig:supp_zero} shows the results of zero-shot generative domain adaption.

\vspace{-0.8em}
\section*{C \quad Zero-shot Generative Domain Adaption}
\vspace{-0.8em}
\label{app_zero_shot}
Compared to one-shot generative domain adaption, we remove the attentive style loss $\Ls_{local}$ and modify the global-level adaption loss $\Ls_{global}$ under the zero-shot setting.
Specifically, we compute the domain-gap direction $\Delta \vv_{dom\_text}$ between the CLIP-space embedding $\vv_{tar\_text}$ of the given text $T_{tar}$ and the text-based embedding $\vv_{src\_text}$ of source domain:
\begin{equation}
    \label{eq:dom_text}
    \Delta \vv_{dom\_text} = \vv_{tar\_text} - \vv_{src\_text},
\end{equation}
where $\vv_{tar\_text} = E_T(T_{tar})$ denotes the embedding of target text $T_{tar}$, and $\vv_{src\_text} = \E_{t_i \in \gX_T}[E_T(t_i)] $ indicates the mean embedding of $N_T$ words $\gX_T=\{t_i\}_{i=1}^{N_T}$ closest to the mean source image embedding $\vv_{\bar A}$.
$E_T$ is the CLIP text encoder.
% where $\vv_{tar\_text} = E_T(T_{tar})$ denotes the embedding of target text $T_{tar}$ and $E_T$ is the CLIP text encoder.
% $\vv_{src\_text} = \E_{t_i \in \gX_T}[E_T(t_i)] $, \ie, the mean embedding of $N_T$ words $\{t_i\}_{i=1}^{N_T}$ closest to the mean source image embedding $\vv_{src}$.
%
Together with the sample-shift direction $\Delta \vv_{samp}$ computed by Eq.~\ref{eq:d_sample}, the text-based global-level adaption loss is defined as:
\begin{equation}
\label{eq:global_text}
    \Ls_{global\_text} = 1 - \frac {\Delta \vv_{samp} \cdot \Delta \vv_{dom\_text}} {|\Delta \vv_{samp}||\Delta \vv_{dom\_text}|},
\end{equation}
Consequently, the overall training loss is expressed as:
\begin{equation}
    \label{eq:final_text}
    \Ls_{overall\_text} = \Ls_{global\_text} + \lambda_{scc}\Ls_{scc}.
\end{equation}
Empirically, $\lambda_{scc}$ is set to four, and we choose $N_T=50$ closest words from the dictionary\footnote{\url{https://github.com/openai/CLIP/blob/main/clip/bpe_simple_vocab_16e6.txt.gz}} of CLIP.
Table~\ref{tab:src_words} shows the chosen words from different source domains.
%
% \vspace{-0.5em}
\section*{D \quad More Ablation Studies}
\label{app_ablation}
\begin{figure}[t]
  \vspace{1em}
  \begin{center}
  \includegraphics[width=.99\linewidth]{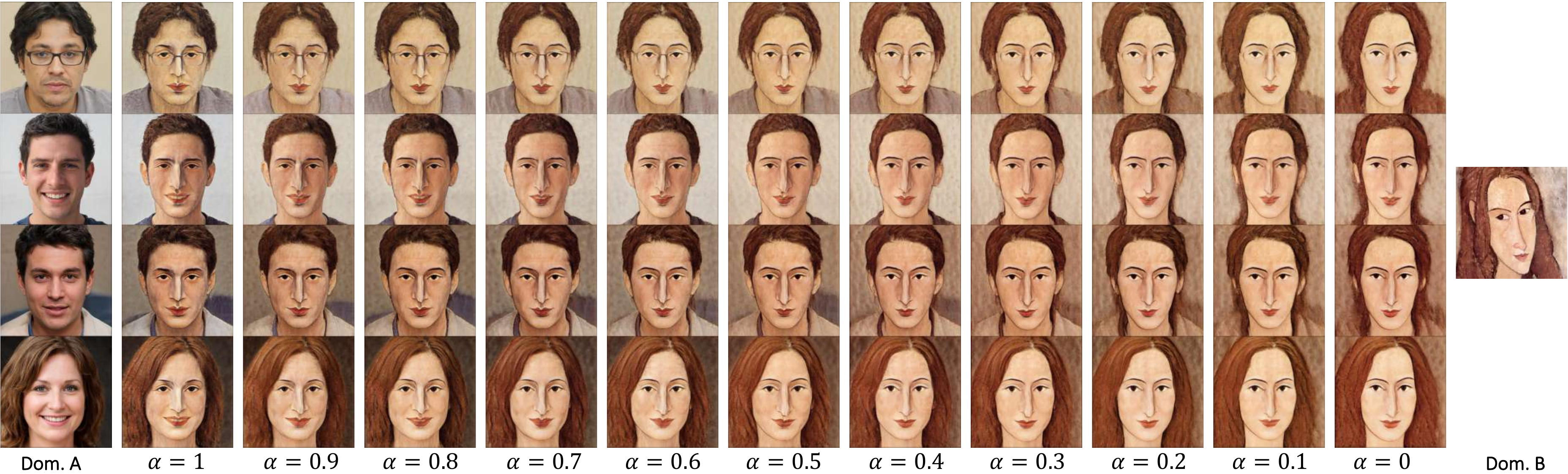}
  \end{center}
  \vspace{-2mm}
  \caption{
  \textbf{Ablation studies on the proportion $\alpha$ of preserved attributes.}
    The first and last column show the source images from domain $A$ and the reference image from domain $B$.
    The other columns show the results using linearly decreasing $\alpha$ from one to zero, \ie, preserving less attributes.
   }
    \label{fig:supp_alpha}
  \vspace{-2.5mm}
\end{figure}

\begin{figure}[t]
   \vspace{-1em}
  \begin{center}
  \includegraphics[width=.99\linewidth]{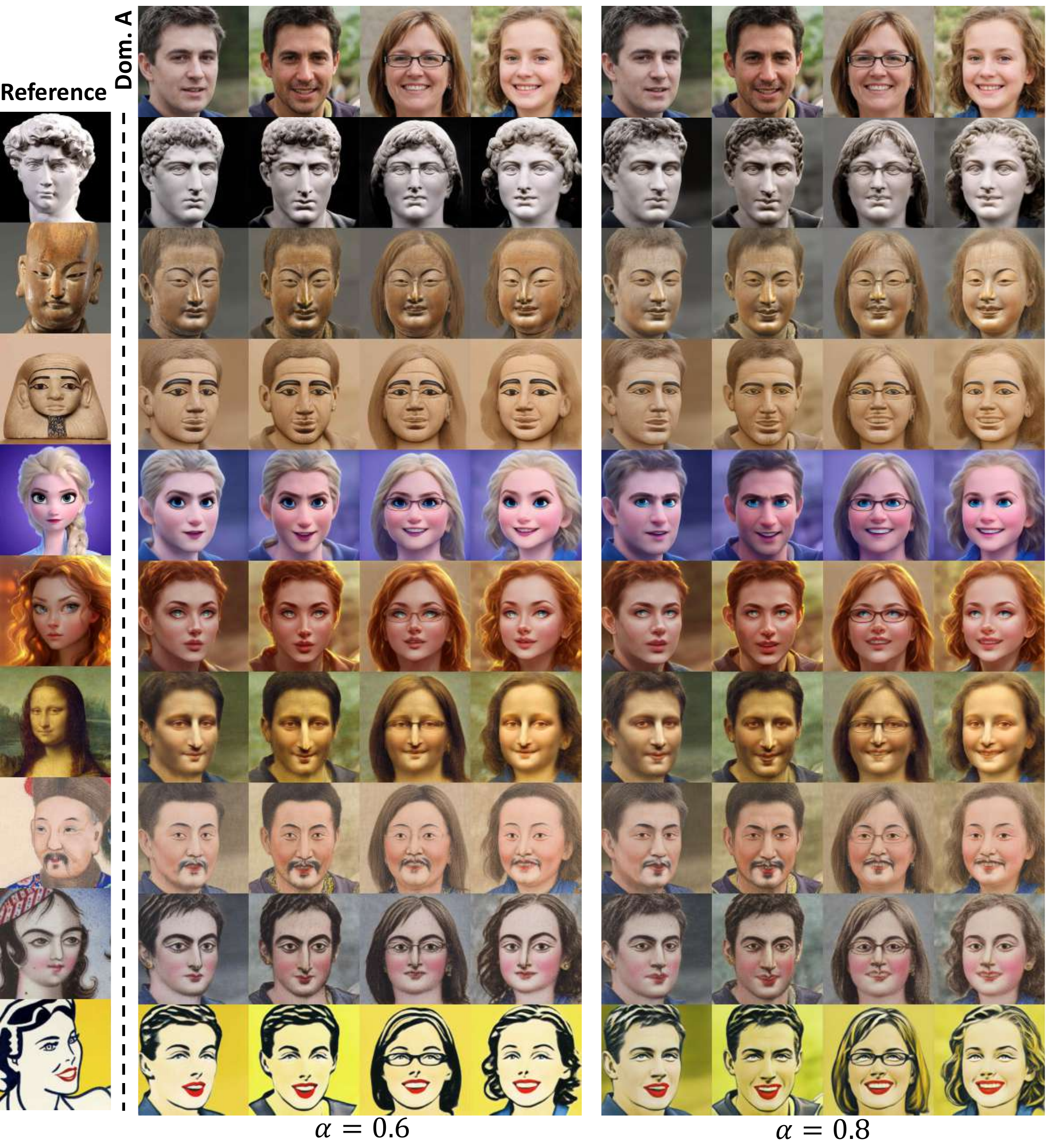}
  \end{center}
  \vspace{-2mm}
  \caption{
\textbf{Qualitative results comparisons $\alpha$=0.6 (original) and $\alpha$=0.8 using the generator pre-trained on FFHQ}.
  The first row and first column show source images in domain $A$ and reference images in domain $B$.
  \textbf{Results best seen at 500\% zoom.}
  }
    \label{fig:larger_ffhq}
  \vspace{-2.5mm}
\end{figure}

\myparagraph{The Proportion $\alpha$ of Preserved Attributes.}
To explore the effect of hyper-parameter $\alpha$ in Eq.~\ref{eq4}, we conduct experiments with $\alpha$ linearly decreasing from one to zero in Fig.~\ref{fig:supp_alpha}.
When decreasing the value of $\alpha$, the synthesized images become more similar to the reference image in the domain-specific attributes (\eg, slim and long face) while having less diversity.
Noticeably, when $\alpha$ is less than 0.5 or greater than 0.7, the adapted generator fails to retain enough domain-sharing attributes (\eg, gender and hair length) or acquire domain-specific attributes (\eg, slim and long face).
Thus, we set $\alpha$=0.5$\sim$0.7 in our experiments.
Notably, as show in Fig.~\ref{fig:larger_ffhq}, a larger $\alpha$ enhances the preservation of face identities for FFHQ source domain.

\begin{figure}[h]
  \vspace{-1em}
  \begin{center}
  \includegraphics[width=.99\linewidth]{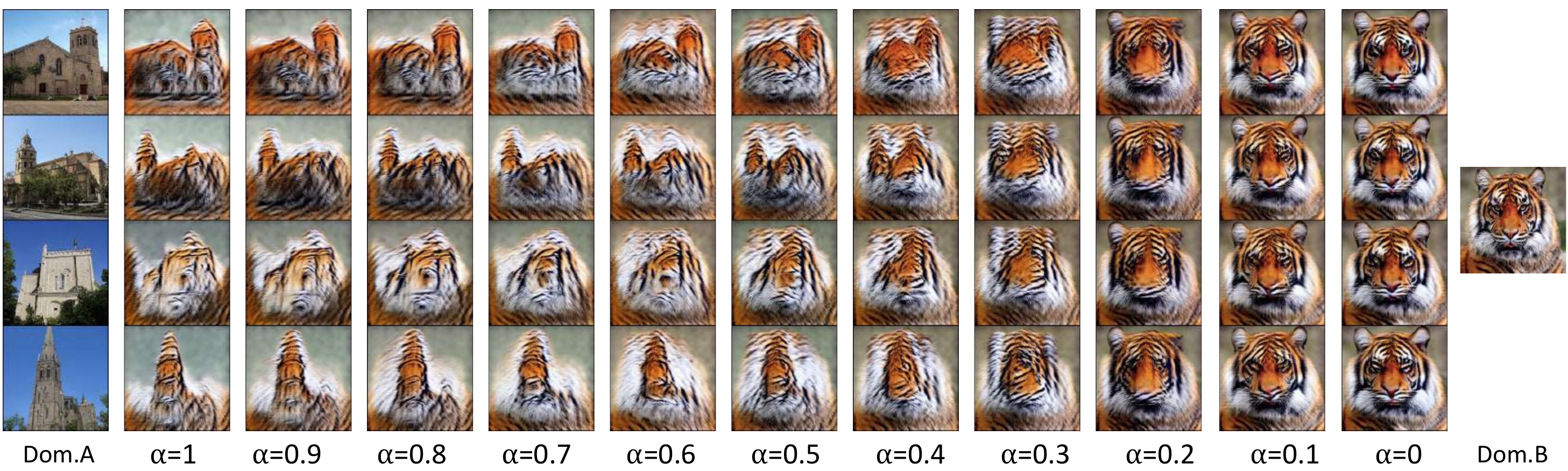}
  \end{center}
  \vspace{-2mm}
  \caption{
\textbf{Ablation studies on the proportion $\alpha$ of preserved attributes for dissimilar domains (church $\to$ tiger).}
    The first and last column show the source images from domain $A$ and the reference image from domain $B$.
    The other columns show the results using linearly decreasing $\alpha$ from one to zero, \ie, preserving less attributes.
   }
    \label{fig:church_tiger}
%   \vspace{-2.5mm}
\end{figure}

In Fig.~\ref{fig:church_tiger}, we also conduct additional ablation studies on the choice of $\alpha$ when the source and target domains are quite dissimilar (\ie, church$\rightarrow$tiger).
Since there are few domain-sharing attributes between church and tiger domains, the selective cross-domain consistency loss favors selecting and retaining the attributes with fewer changes (\eg, shape and pose).
Specifically, when the value of $\alpha$ is very large (0.7$\sim$1), the synthesized images keep the shape of "church" while acquiring the fur and stripes of "tiger". After linearly decreasing the value of $\alpha$, we observe that the adapted generator produces images with coarser-scale characteristics of the target domain (\eg, the face shape of the tiger). Until $\alpha$ is decreased to zero, all adapted images look very similar to the reference image with little diversity.

\begin{figure}[t]
%   \vspace{-1em}
  \begin{center}
  \includegraphics[width=.99\linewidth]{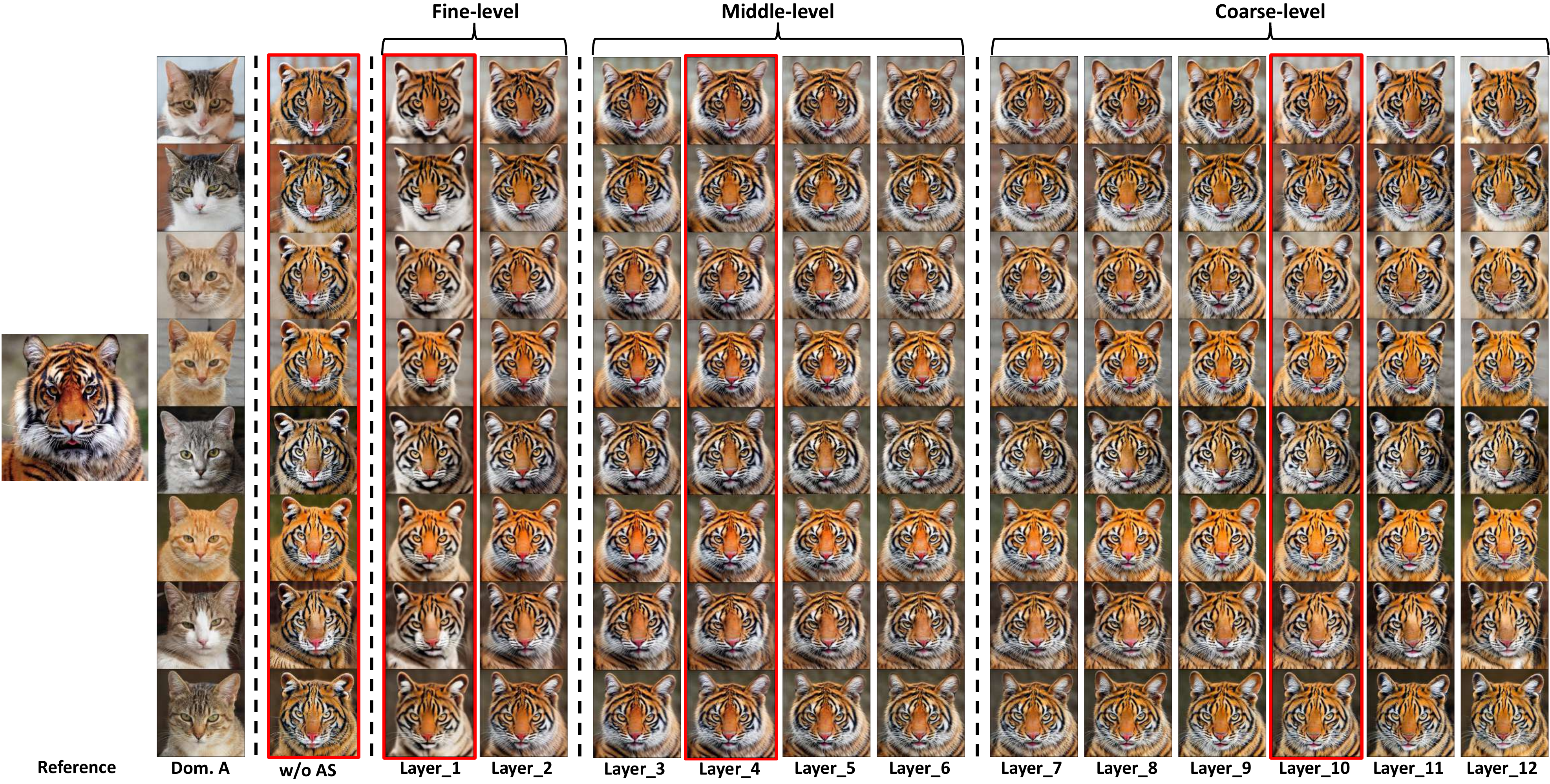}
  \end{center}
%   \vspace{-2mm}
  \caption{
  \textbf{Ablation studies on the layer choice of attentive style (AS) loss.}
    The first two column show the reference image from domain $B$ and the source images from domain $A$.
    The other columns show the results under the different AS configurations, \ie, using the intermediate tokens from different layers of CLIP image encoder.
    Akin to ~\cite{johnson2016perceptual}, we divide all layers into fine-level, middle-level, and coarse-level, and then select a layer (in \textcolor{red}{red} box) from each stage for better comparison.
  }
    \label{fig:supp_layer}
  \vspace{5mm}
\end{figure}

\myparagraph{Layer Choice of Attentive Style Loss.}
We also investigate the effect of different layer choices on the performance of attentive (AS) style loss.
In Fig.~\ref{fig:supp_layer}, all intermediate layers are divide into fine-level (1-2), middle-level (3-6), and coarse-level (7-12).
We can observe that fine-level layers only capture fine-grained characteristics (\eg, fur color) of the reference image.
Coarse-level layers obtain similar performance with the one w/o AS, because intermediate tokens become more similar to the final CLIP-space embedding as the layers deepen.
In contrast, middle-level layers acquire both representative domain styles (\eg, fur color and stripes) and attributes (\eg, mane), hence, we use intermediate tokens from the $4$-th layer of CLIP image encoder by default.

\myparagraph{Quantitative Ablation Studies of Proposed Losses.}
We have added quantitative ablation studies on the effectiveness of our proposed two losses in Table~\ref{tab:ablation_kid} and ~\ref{tab:ablation_fid}. 
From the tables, both selective cross-domain consistency loss and attentive style losses can boost the performance of one-shot domain adaption in terms of KID and FID scores, which is consistent with the qualitative ablation studies.

\section*{E \quad More Comparison Experiments}
\label{app_comparison}
\begin{figure}[t]
%   \vspace{-1em}
  \begin{center}
  \includegraphics[width=.99\linewidth]{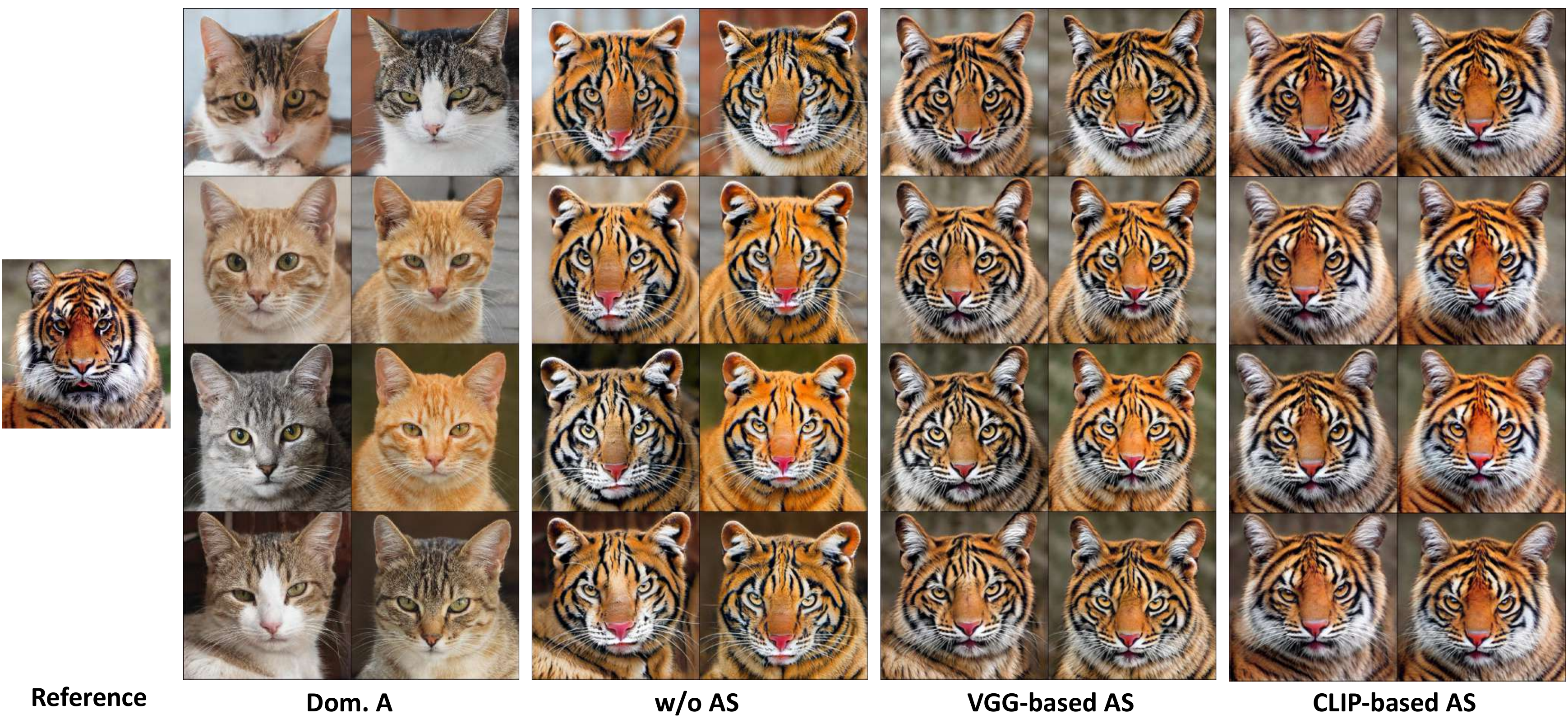}
  \end{center}
%   \vspace{-2mm}
  \caption{
  \textbf{CLIP-based vs. VGG-based attentive style (AS) loss.}
    The first two columns show the reference from domain $B$ and the source images from domain $A$ and the reference images from domain $B$, respectively.
    Akin to ~\cite{kolkin2019style}, VGG-based AS uses all layers of VGG16 but layers 9, 10, 12, and 13.
    In contrast, our CLIP-based AS only uses the $4$-th layer of CLIP image encoder.
  }
    \label{fig:supp_vgg}
  \vspace{5em}
\end{figure}

\myparagraph{CLIP-based AS vs VGG-based AS.}
Fig.~\ref{fig:supp_vgg} shows comparisons between CLIP-based and VGG-based attentive style (AS) loss.
As can be seen, VGG-based AS only captures some visual styles (\eg, stripes) from the reference, while CLIP-based AS acquires more representative domain styles (\eg, fur color and stripes) and attributes (\eg, mane), even using one intermediate layer only.

\begin{figure}[t]
   \vspace{-1em}
  \begin{center}
  \includegraphics[width=.99\linewidth]{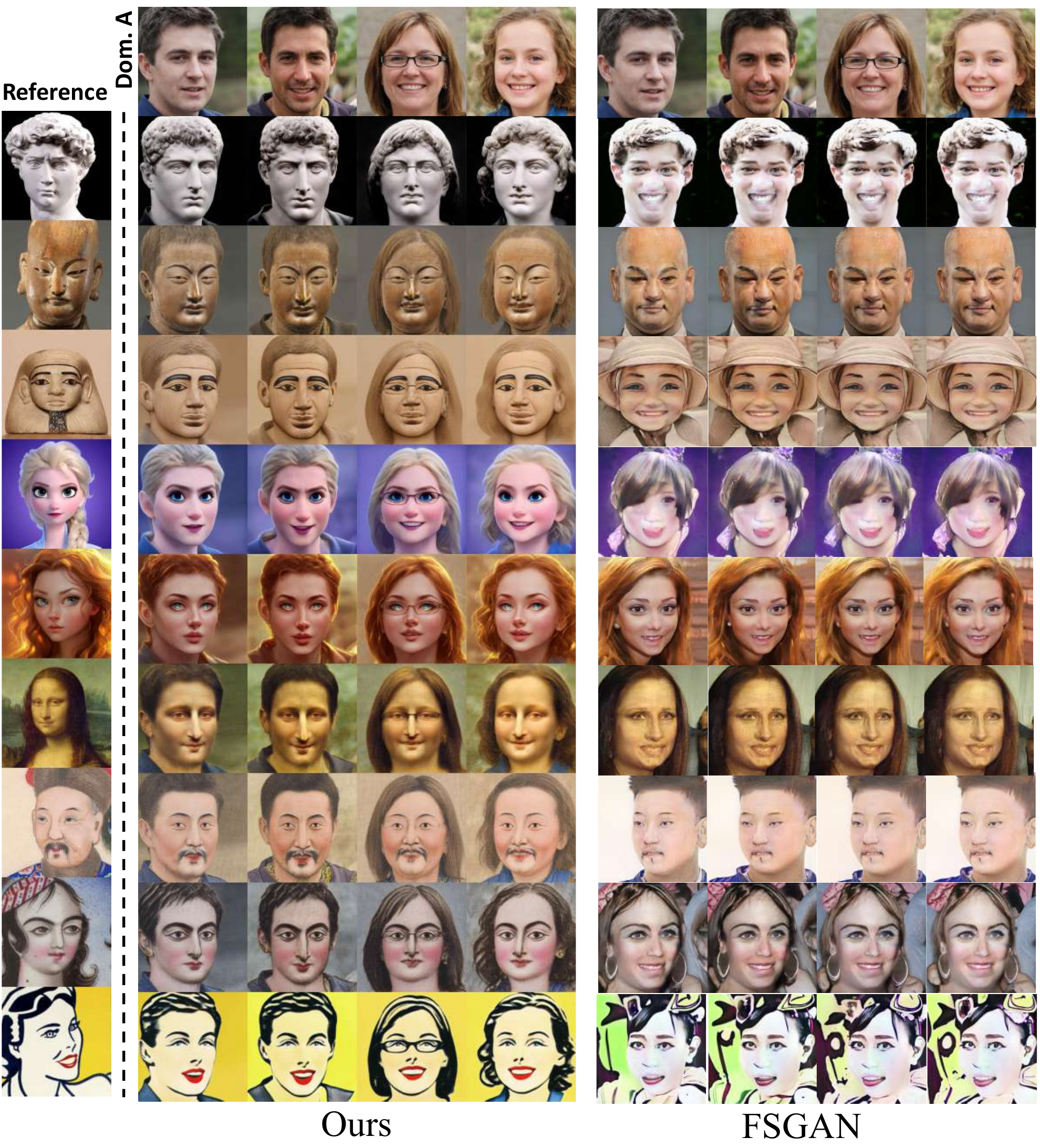}
  \end{center}
  \vspace{-2mm}
  \caption{
\textbf{Qualitative comparisons using the generator pre-trained on FFHQ}~\cite{karras2019style} between our DiFa, and FSGAN~\cite{robb2020few}.
  The first row and first column show source images in domain $A$ and reference images in domain $B$.
  In contrast, FSGAN suffers from severe mode collapse and fails to obtain domain-specific styles of the reference images.
  \textbf{Results best seen at 500\% zoom.}
  }
    \label{fig:compare_adv}
  \vspace{-2.5mm}
\end{figure}

\myparagraph{Our DiFa vs Adversarial Loss Methods.}
We present the qualitative comparisons with FSGAN~\cite{robb2020few} in Fig.~\ref{fig:compare_adv} and quantitative comparisons with FSGAN~\cite{robb2020few} and GenDA~\cite{yang2021one} in Table~\ref{tab:adv_kid} and ~\ref{tab:adv_fid}. 
As shown in Fig.~\ref{fig:compare_adv}, FSGAN~\cite{robb2020few} not only suffers from severe mode collapse but also fails to capture domain-specific styles of the reference images. 
In terms of quantitative results, our DiFa significantly outperforms the methods based on adversarial loss by the KID and FID metrics under the one-shot setting, which is consistent with qualitative results.

\begin{figure}[t]
  \vspace{-1em}
  \begin{center}
  \includegraphics[width=.99\linewidth]{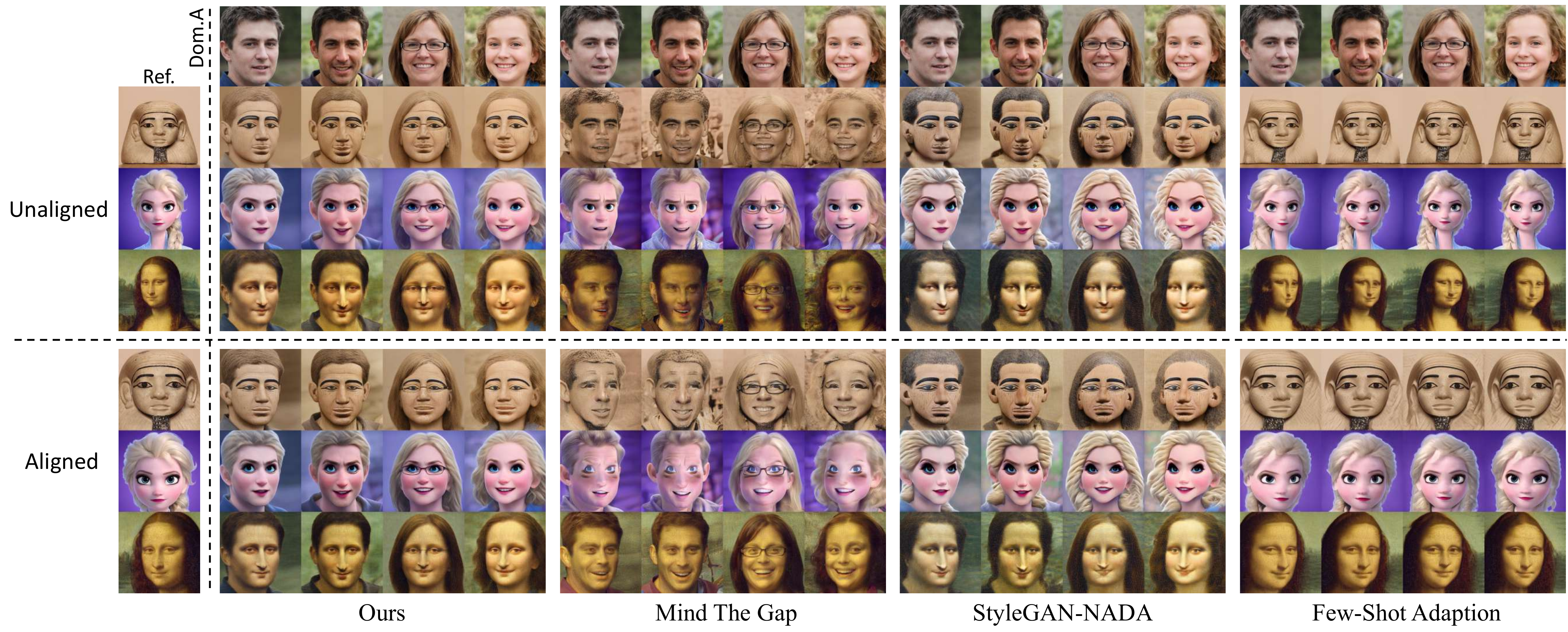}
  \end{center}
  \vspace{-2mm}
  \caption{
\textbf{Qualitative results using unaligned and aligned reference images.}
    The first row and column show the source images from domain $A$ and the reference image from domain $B$.
   }
    \label{fig:aligned_ffhq}
  \vspace{-2.5mm}
\end{figure}

\begin{figure}[h!]
  \vspace{-1em}
  \begin{center}
  \includegraphics[width=.99\linewidth]{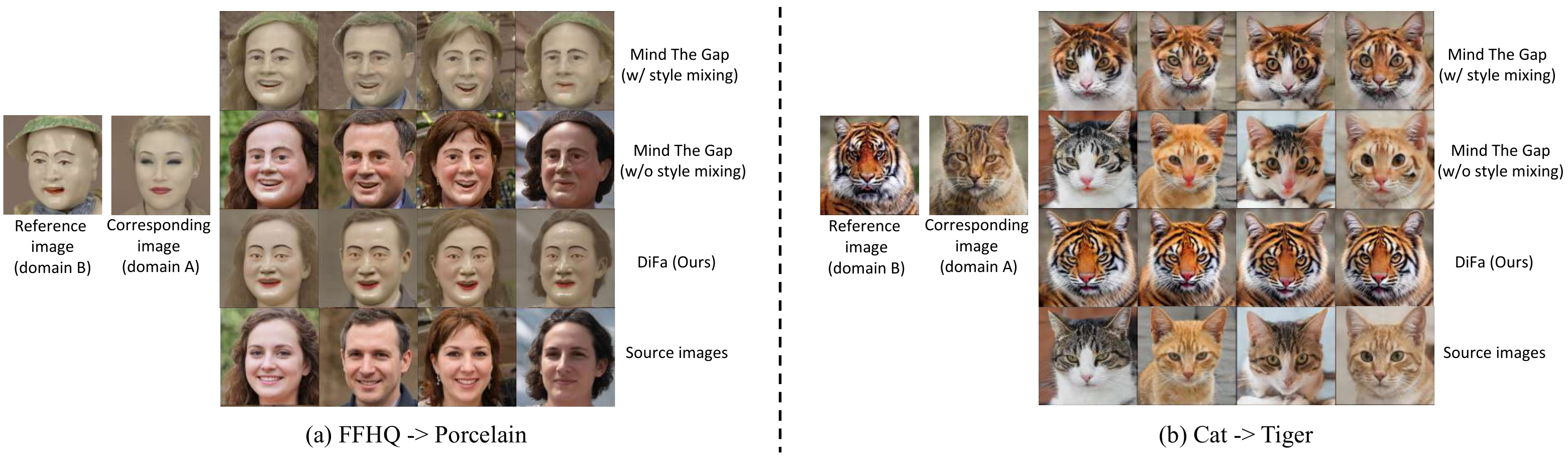}
  \end{center}
  \vspace{-2mm}
  \caption{
\textbf{Qualitative comparisons between attentive loss based (our DiFa) and style mixing based method (Mind The Gap~\cite{zhu2021mind})}.
  For both (a) and (b), the first and second columns show reference images in domain $B$ and their corresponding images in domain $A$ respectively.
  The first row shows source images in domain $A$.
%   Mind The Gap finds the corresponding image in domain A of the reference image and takes its CLIP embedding as source domain embedding, thereby ignoring some domain-specific characteristics (\eg, glaze color and red lips in (a), fur color and mane of tiger in (b)).
%   After that, it tries to re-acquire ignored domain-specific characteristics using the style mixing. Yet, it still fails to capture some representative attributes (\eg, red lips in row 1 of (a), mane and stripes in row 1 of (b)) and introduces visible artifacts (\eg, mistake the green hat as hair in row 1).
  }
    \label{fig:style_mixing}
%   \vspace{-2.5mm}
\end{figure}

\myparagraph{Attentive Style Loss vs Style Mixing}
With the aid of intermediate tokens of CLIP model, our attentive style loss directly encourages the model to learn to acquire the target styles. 
While style mixing acquires the target styles through the obtained latent code of the reference image, and the style heavily relies on the latent code. Usually, it is difficult to faithfully obtain the latent code of the reference image, especially for the images with rare or unseen attributes for the source domain.
Therefore, our attentive style loss is more robust than the style mixing trick when dealing with cases involving a large domain gap, \eg, Cat$\rightarrow$Tiger in Fig.~\ref{fig:style_mixing}(b)) and using unaligned reference images in Fig.~\ref{fig:aligned_ffhq}.

\begin{figure}[t]
   \vspace{-1em}
  \begin{center}
  \includegraphics[width=.99\linewidth]{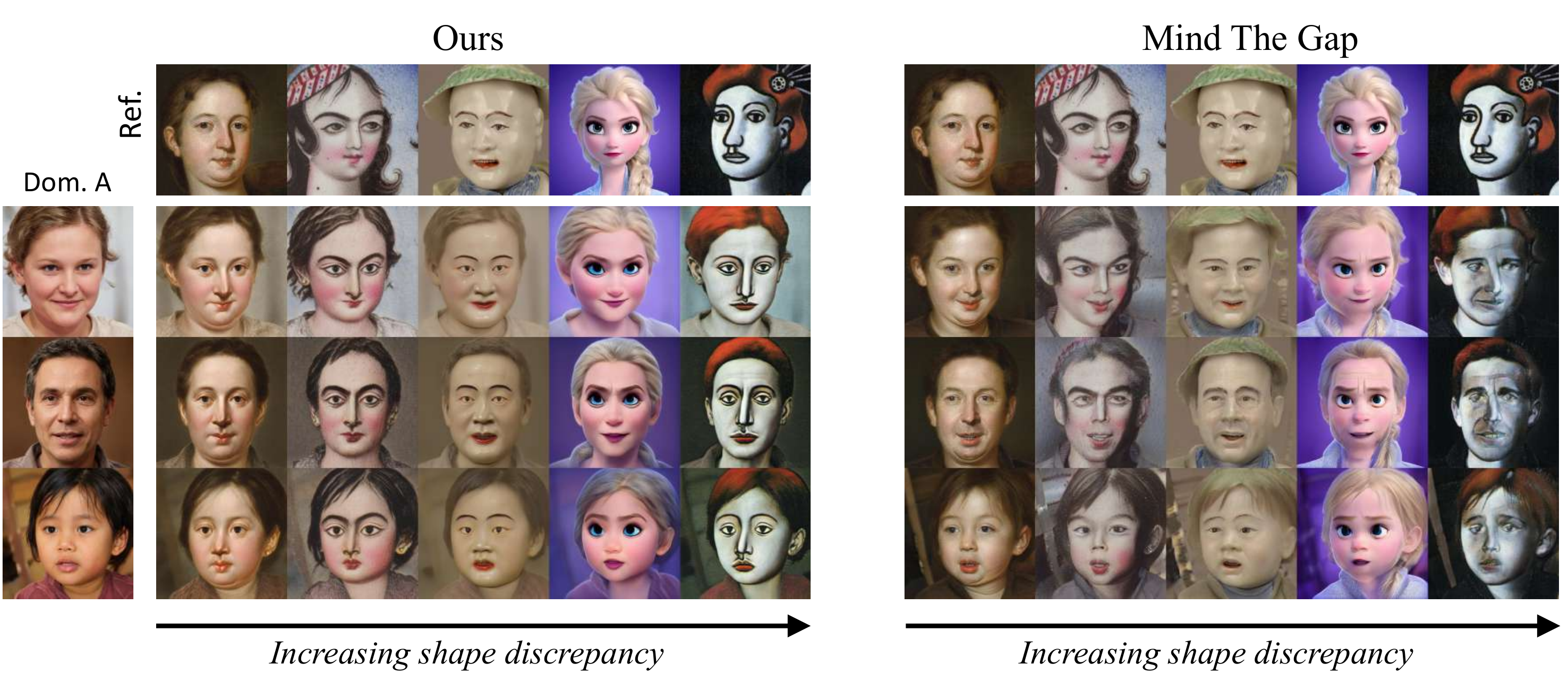}
  \end{center}
  \vspace{-2mm}
  \caption{
\textbf{Qualitative comparisons when increasing the shape discrepancy between source images and the reference image} between our DiFa and Mind The Gap~\cite{zhu2021mind}.
  The first row and first column show reference images in domain $B$ and source images in domain $A$.
  The shape discrepancy between a reference image and a source image is defined as the L2 normalized distance between their face landmarks.
  When increasing the shape discrepancy from left to right, our DiFa produces images with high quality and diversity while Mind The Gap produces images with more visible artifacts.
  }
    \label{fig:shape_dis}
  \vspace{-2.5mm}
\end{figure}

To quantitatively evaluate the shape discrepancy of faces, we calculate the distances between the landmarks of two different faces. 
In particular, we use the dlib library\footnote{\url{http://dlib.net/face_landmark_detection.py.html}} to detect 68 landmarks of the human face and take the Euclidean distance between landmarks of a reference image and a source image as their shape discrepancy. 
Fig.~\ref{fig:shape_dis} illustrates the comparison between the style mixing method and our DiFa as the increase of the shape discrepancy.
As one can see, the style mixing method synthesizes images with more visible artifacts when increasing the shape discrepancy. 
In contrast, our DiFa is minimally affected by the shape discrepancy and keeps producing images with high quality and diversity.

\myparagraph{One-stage vs Two-stage Methods.}
In Fig.~\ref{fig:style_mixing}, we present the results of a two-stage method (Mind The Gap~\cite{zhu2021mind}). 
One can see that the two-stage method ignores some domain-specific attributes (\eg, red lips in row 1 of Fig.~\ref{fig:style_mixing}(a), manes and stripes in row 1 of Fig.~\ref{fig:style_mixing}(b)), even using the style mixing trick during inference. 
Specifically, the two-stage method finds the corresponding image in source domain of the reference image and treats its CLIP embedding as source domain embedding. 
As shown in Fig.~\ref{fig:style_mixing}, the found corresponding image contains some domain-specific attributes of the reference image (\eg, glaze color and red lips in Fig.~\ref{fig:style_mixing}(a)). 
And the domain gap on these attributes is negligible, thereby ignoring these domain-specific attributes during adaption. 
Albeit the two-stage method tries to re-acquire ignored domain-specific attributes using style mixing, it still fails to acquire some of them (\eg, red lips in row 1 of Fig.~\ref{fig:style_mixing}) or misunderstands some attributes (e.g., mistake the green hat as green hair).

\section*{F \quad User Study}
\begin{figure}[t!]
   \begin{center}
   \includegraphics[width=.99\linewidth]{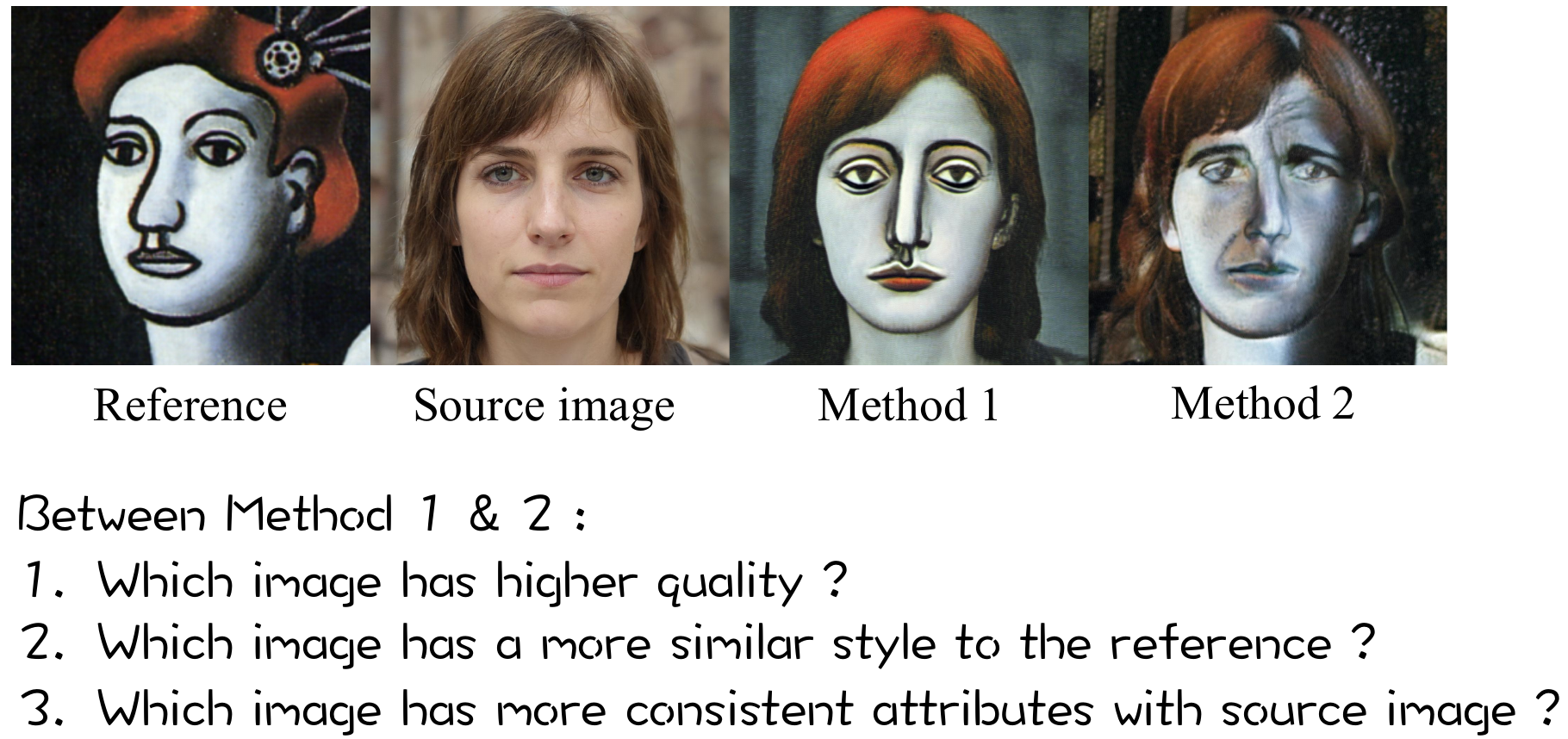}
   \end{center}
  \vspace{-2mm}
   \caption{
   \textbf{Instructions of user study.}
   A user study sample consists of a reference, a source image, and two adapted images from our DiFa and another method.
   The participates are asked to answer the above three questions for each sample.
   Note that two adapted images are randomly permuted to reduce potential position bias.
    }
    \label{fig:user_instructions}
    \vspace{-1em}
\end{figure}

We perform user study to further compare our DiFa with other approaches, from the perspective of (i) image quality, (ii) style similarity and (iii) attribute consistency.
We recruit 30 participates from both universities and industries, whose statistics are shown in Table~\ref{tab:user_stat}.
Particularly, we randomly generate 1,050 samples for each "our DiFa vs another method" comparison.
Afterwards, we assign these samples to 30 participates and ask them to complete the survey following the instructions in Fig.~\ref{fig:user_instructions}.
Finally, we collect their answers and illustrate the statistics in Table~\ref{tab:main_user}.

\begin{table}[t]
\vspace{-1em}
\caption{
\textbf{The 50 words closest to the mean embedding $\vv_{\bar A}$ of source images on different source domains.}
}
\vspace{1mm}
\centering
\scalebox{0.8}{
\begin{tabular}{l|c}
\toprule
\makecell{\textbf{Source Domain}} & \makecell{\textbf{Chosen Words}} \\
\midrule
FFHQ & \makecell{``person'', ``headshot'', ``participant'', ``face'', ``closeup'', ``filmmaker'', ``author'',\\
 ``pknot'', ``contestant'', ``associate'', ``individu'', ``volunteer'', ``michele'',\\
 ``artist'', ``director'', ``researcher'', ``cropped'', ``lookalike'', ``mozam'', ``ml'',\\
 ``portrait'', ``organizer'', ``kaj'', ``coordinator'', ``appearance'', ``psychologist'',\\
 ``jha'', ``pupils'', ``subject'', ``entrata'', ``newprofile'', ``guterres'', ``staffer'',\\
 ``diem'', ``cosmetic'', ``viewer'', ``assistant'', ``writer'', ``practitioner'',\\
 ``adolescent'', ``white'', ``elling'', ``nikk'', ``addic'', ``onnell'', ``customer'', ``client'',\\
 ``simone'', ``greener'', ``candidate''} \\
\midrule
AFHQ-Cat & \makecell{``burmese'', ``feline'', ``cat'', ``tabby'', ``cathedr'', ``gata'', ``gato'', ``alcat'', ``catt'', \\
 ``pupils'', ``wildcat'', ``bengal'', ``tuna'', ``artemis'', ``feral'', ``persian'', ``meow'', \\
 ``figaro'', ``packet'', ``java'', ``cappuccino'', ``tora'', ``alley'', ``catal'', ``chipped'', \\
 ``kitty'', ``cathar'', ``miaw'', ``pye'', ``chattanoo'', ``katz'', ``sniff'', ``kerswednesday'', \\
 ``peuge'', ``categor'', ``nak'', ``mae'', ``catalo'', ``scratch'', ``tabern'', ``plume'', \\
 ``striped'', ``chat'', ``catsofinstagram'', ``cajun'', ``meredith'', ``offee'', ``sylvester'', \\
 ``popart'', ``pling''} \\
\midrule
AFHQ-Dog & \makecell{``adog'', ``dog'', ``canine'', ``adoptable'', ``doggie'', ``doggy'', ``mutt'', ``pupp'',\\
``doggo'', ``terrier'', ``pup'', ``cajun'', ``pooch'', ``dogday'', ``maverick'', ``dawg'', ``watchdog'',\\
``lostdog'', ``peuge'', ``woof'', ``skye'', ``tucker'', ``sampson'', ``detect'', ``dug'', ``kodi'', ``embark'',\\
``renegade'', ``puppy'', ``wrangler'', ``hula'', ``ruff'', ``sabre'', ``zeus'', ``dharma'', ``wag'', ``cooper'',\\
``brownie'', ``aviator'', ``kita'', ``bud'', ``cigar'', ``shepherd'', ``chaser'', ``dixie'', ``taro'', ``scotch'',\\
``duke'', ``tobi'', ``bullet''} \\
\midrule
LSUN CAR & \makecell{``car'', ``ecar'', ``vehicle'', ``automobile'', ``automotive'', ``autonews'', ``icar'',\\
``hatchback'', ``incar'', ``auto'', ``saab'', ``sedan'', ``lowered'', ``classiccar'', ``cars'', ``nissan'',\\
``stance'', ``citroen'', ``supercharged'', ``tuned'', ``facelift'', ``volvo'', ``convertible'', ``forza'', \\
``skoda'', ``civic'', ``sportscar'', ``oem'', ``opel'', ``mazda'', ``valet'', ``extravag'', ``merc'', ``].'',\\
``parked'', ``chevrolet'', ``suv'', ``coupe'', ``slammed'', ``xf'', ``toyota'', ``gtx'', ``bmw'', ``corsa'',\\
``tdi'', ``taxi'', ``amg'', ``detailing'', ``peugeot'', ``spoiler''} \\
\midrule
LSUN CHURCH & \makecell{``cathedral'', ``church'', ``churches'', ``basilica'', ``chapel'', ``anglican'', ``lutheran'',\\
``diocese'', ``dral'', ``presbyterian'', ``apse'', ``friars'', ``st'', ``cathol'', ``methodist'', ``abbey'', \\
``baptist'', ``synagogue'', ``conduc'', ``argu'', ``assumption'', ``jesu'', ``congregation'', ``priory'', ``nave'',\\
``episcopal'', ``halle'', ``exterior'', ``cst'', ``sedly'', ``echel'', ``mably'', ``gonzaga'', ``nd'', ``protestant'',\\
``thex'', ``monastery'', ``cour'', ``bishops'', ``mons'', ``minster'', ``tor'', ``sacrific'', ``shul'', ``heritag'',\\
``sque'', ``restoration'', ``wul'', ``spires'', ``notre''} \\
\bottomrule
\end{tabular}
}
\label{tab:src_words}
\vspace{-2mm}
\end{table}

\begin{table}[t]
% \vspace{-1em}
\caption{
\textbf{Statistics of participates in user study.}
}
\vspace{1mm}
\centering
\scalebox{1}{
\begin{tabular}{l|c}
\toprule
\makecell{\textbf{Factors}} & \makecell{\textbf{Statistics}} \\
\midrule
Gender  &{Male: 53.3$\%$, Female: 47.7$\%$} \\
Age  &{$\leq$20: 23.3$\%$, 20$\sim$40: 56.7$\%$, $\ge$ 40: 20$\%$} \\
Background &{CV and CG: 33.3$\%$, Arts: 36.7$\%$, Other$\%$} \\
Race &{Caucasian: 26.7$\%$, Mongoloid: 33.3$\%$, Negroid: 23.3$\%$, Australoid: 16.7$\%$}\\
\bottomrule
\end{tabular}
}
\label{tab:user_stat}
\vspace{-4mm}
\end{table}

\begin{table}[t]
   \centering
   \caption{\textbf{KID ($\downarrow$) of ablation studies on selective cross-domain consistency (SCC) and attentive style (AS) losses.} Each result is averaged over 5 training shots and in the form of $\{$mean $\pm$ standard error$\}$.}
%   \vspace{1em}
   \resizebox{0.75\linewidth}{!}{
   \begin{tabular}{cc|ccc}
      \toprule
    %   \cline{4-11}
      \multicolumn{1}{c}{w/ SCC} & \multicolumn{1}{c|}{w/ AS} &
      Amedeo. & Fernand. & Raphael \\
      \midrule
    %   \xmark &\xmark &\xmark &80.96 &84.10 &76.29 &68.31 & 73.58 & 59.65 & 69.55 & 69.02  \\
      {} &{} &131.03$\pm$28.14 &169.83$\pm$ 31.52 &149.19$\pm$ 55.91\\
      \checkmark &{} & 129.41$\pm$ 26.17 & 165.73$\pm$ 31.75 & 119.36$\pm$ 20.18 \\
      \checkmark &\checkmark &\textbf{121.21$\pm$24.62} &\textbf{159.93$\pm$31.39} &\textbf{112.72$\pm$17.61}	\\
      \bottomrule
   \end{tabular}
  }
    \label{tab:ablation_kid}
    % \vspace{-1em}
 \end{table}
\begin{table}[h]
   \centering
   \caption{\textbf{FID ($\downarrow$) of ablation studies on selective cross-domain consistency (SCC) and attentive style (AS) losses.} Each result is averaged over 5 training shots and in the form of \{mean $\pm$ standard error\}.}
   \resizebox{0.75\linewidth}{!}{
   \begin{tabular}{cc|ccc}
      \toprule
    %   \cline{4-11}
      \multicolumn{1}{c}{w/ SCC} & \multicolumn{1}{c|}{w/ AS} &
      Amedeo. & Fernand. & Raphael \\
      \midrule
    %   \xmark &\xmark &\xmark &80.96 &84.10 &76.29 &68.31 & 73.58 & 59.65 & 69.55 & 69.02  \\
      {} &{} &188.44$\pm$19.15 &257.27$\pm$19.39 &186.20$\pm$28.60  \\
      \checkmark &{} & 187.42$\pm$19.32 & 257.18$\pm$21.32 & 180.61$\pm$15.32 \\
      \checkmark &\checkmark &\textbf{187.28$\pm$24.45} &\textbf{254.68$\pm$17.73}  &\textbf{172.34$\pm$10.15}	\\
      \bottomrule
   \end{tabular}
   }
    \label{tab:ablation_fid}
    % \vspace{-1em}
 \end{table}
\begin{table}[h]
% \vspace{-1em}
\caption{
\textbf{KID ($\downarrow$) comparisons between our DiFa and adversarial loss based methods.}
Each result is averaged over 5 training shots and in the form of \{mean $\pm$ standard error\}.}
\vspace{1mm}
\centering
\scalebox{0.95}{
\begin{tabular}{l|cccc}
\toprule
{Models} & {Amedeo.} & {Fernand.} & {Raphael} & {Sketches} \\
\midrule
FSGAN~\cite{robb2020few} &299.64$\pm$ 34.16  &348.70$\pm$ 41.27   &151.79$\pm$ 35.12 &227.78$\pm$12.71\\

\midrule
\textbf{Ours} &\textbf{121.21$\pm$ 24.62} &\textbf{159.93$\pm$31.39} &\textbf{112.72$\pm$17.61} &\textbf{53.24$\pm$7.82}\\
\bottomrule
\end{tabular}
}
\label{tab:adv_kid}
% \vspace{-10em}
\end{table}
\begin{table}[h!]
% \vspace{-10em}
\caption{
\textbf{FID ($\downarrow$) comparisons between our DiFa and adversarial loss based methods.}
Each result is averaged over 5 training shots and in the form of \{mean $\pm$ standard error\}.
* indicates that results are from the original paper.}
\vspace{1mm}
\centering
\scalebox{0.95}{
\begin{tabular}{l|cccc}
\toprule
{Models} & {Amedeo.} & {Fernand.} & {Raphael} & {Sketches} \\
\midrule
FSGAN~\cite{robb2020few} &288.75$\pm$46.76  &360.45$\pm$58.07  &200.29$\pm$78.23  &166.37$\pm$11.32\\
GenDA*~\cite{yang2021one} &- &- &- &87.55 \\
\midrule
\textbf{Ours} \textbf{Ours} &\textbf{187.28$\pm$24.45} &\textbf{254.68$\pm$17.73}  &\textbf{172.34$\pm$10.15} &\textbf{56.93$\pm$5.48}\\
\bottomrule
\end{tabular}
}
\label{tab:adv_fid}
% \vspace{-4mm}
\end{table}


% More Visualizations
% \clearpage

\end{appendices}
\end{document}